\documentclass[journal]{IEEEtran}

\usepackage[linesnumbered,ruled]{algorithm2e}
\usepackage{amsfonts}
\usepackage{amssymb}
\usepackage{array}
\usepackage{bm}
\usepackage{booktabs}
\usepackage{cite}
\usepackage{color}
\usepackage{xcolor}
\usepackage{colortbl, booktabs}
\usepackage{dsfont}
\usepackage{epsfig}
\usepackage{float}
\usepackage{graphicx}
\usepackage{hhline}
\usepackage{lineno}
\usepackage{mathtools}
\usepackage{multirow}
\usepackage{subfigure}
\usepackage{times}
\usepackage[colorlinks,linkcolor=blue]{hyperref}

\graphicspath{{./authors/}}

\begin{document}
\title{DM-CFO: A Diffusion Model for Compositional 3D Tooth Generation with Collision-Free Optimization}
\author{
Yan Tian, 
Pengcheng Xue,
Weiping Ding, \IEEEmembership{Senior Member, IEEE}, 
Mahmoud Hassaballah, \IEEEmembership{Member, IEEE},
Karen Egiazarian, \IEEEmembership{Fellow, IEEE}, 
Aura Conci,
Abdulkadir Sengur,
Leszek Rutkowski, \IEEEmembership{Life Fellow, IEEE}
\thanks{Manuscript received September 9, 2025; revised December 22, 2025.
This work was supported by the Zhejiang Province Natural Science Foundation (No. LZ24F020001), AGH University of Krakow "Excellence initiative - research university”, Polish Ministry of Science and Higher Education funds (No. UMO-2021/01/2/ST6/00004 and No. ARTIQ/0004/2021), the Opening Foundation of the Tongxiang Institute of General Artificial Intelligence (No. TAGI2-B-2024-0009), and the State Key Laboratory of Advanced Medical Materials and Devices (No. SQ2022SKL01089-2025-14).}
\thanks{Yan Tian and Pengcheng Xue are with the School of Computer Science and Technology, Zhejiang Gongshang University, Hangzhou 310018, China. Yan Tian also works for Shining3D Tech Co., Ltd., Hangzhou 311200, China.}
\thanks{Weiping Ding is with the School of Artificial Intelligence and Computer Science, Nantong University, Nantong 226019, China.}
\thanks{Mahmoud Hassaballah is with the Department of Computer Science, College of Computer Engineering and Sciences, Prince Sattam Bin Abdulaziz University, AlKharj, 16278, Saudi Arabia. He also is with the Department of Computer Science, Qena University, Qena 83523, Egypt.} 
\thanks{Karen Egiazarian is with the Department of Computing Sciences, Tampere University, Tampere 33720, Finland.}
\thanks{Aura Conci is with the Department of Computer Science, Universidade
Federal Fluminense, Niteroi 24210-346, Brazil.}
\thanks{Abdulkadir Sengur is with the Department of Electrical and Electronic
Engineering, Faculty of Technology, Firat University, 23000 Elazig, Turkey.}
\thanks{Leszek Rutkowski is with the Systems Research Institute of the Polish Academy of Sciences, 01-447 Warsaw, Poland, with AGH University of Krakow, 30-059 Krakow, and with the SAN University, 90-113 Lodz, Poland.}
\thanks{Corresponding author: Yan Tian and Weiping Ding, 
email: tianyan@zjgsu.edu.cn, dwp9988@163.com.}}

\markboth{IEEE Transactions on Visualization and Computer Graphics,~Vol.~41, No.~8, August~2026}%
{Shell \MakeLowercase{\textit{et al.}}: Bare Demo of IEEEtran.cls for IEEE Journals}
%

\maketitle
\begin{abstract}
The automatic design of a 3D tooth model plays a crucial role in dental digitization. However, current approaches face challenges in compositional 3D tooth generation because both the layouts and shapes of missing teeth need to be optimized.
In addition, collision conflicts are often omitted in 3D Gaussian–based compositional 3D generation, where objects may intersect with each other due to the absence of explicit geometric information on the object surfaces.
Motivated by graph generation through diffusion models and collision detection using 3D Gaussians, we propose an approach named DM-CFO for compositional tooth generation, where the layout of missing teeth is progressively restored during the denoising phase under both text and graph constraints.
Then, the Gaussian parameters of each layout-guided tooth and the entire jaw are alternately updated using score distillation sampling (SDS).
Furthermore, a regularization term based on the distances between the 3D Gaussians of neighboring teeth and the anchor tooth is introduced to penalize tooth intersections.
Experimental results on three tooth-design datasets demonstrate that our approach significantly improves the multiview consistency and realism of the generated teeth compared with existing methods.
Project page: \href{https://amateurc.github.io/CF-3DTeeth/}{https://amateurc.github.io/CF-3DTeeth/}
\end{abstract}


\section{Introduction}
\IEEEPARstart{T}{ooth} model design, a specialized area within the medical applications of computer graphics, involves the creation of accurate and detailed representations of teeth for various dental applications, including prosthetics, orthodontics, and restorative dentistry~\cite{li2023imtooth,dastan2024co}. This design process is crucial to ensure that dental restorations are properly fitted and function effectively. The field has been notably advanced through the integration of computer-aided design (CAD) technology, as well as recent developments in deep learning techniques.

\begin{figure}[htbp]
\centering
\subfigure[GALA3D]{\includegraphics[width=0.98\linewidth]{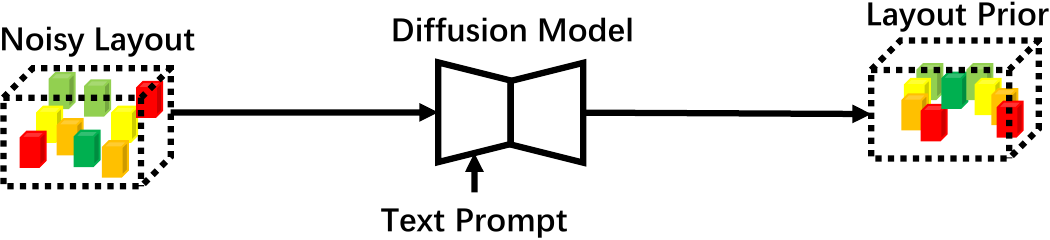}}\\
\subfigure[DM-CFO]{\includegraphics[width=0.98\linewidth]{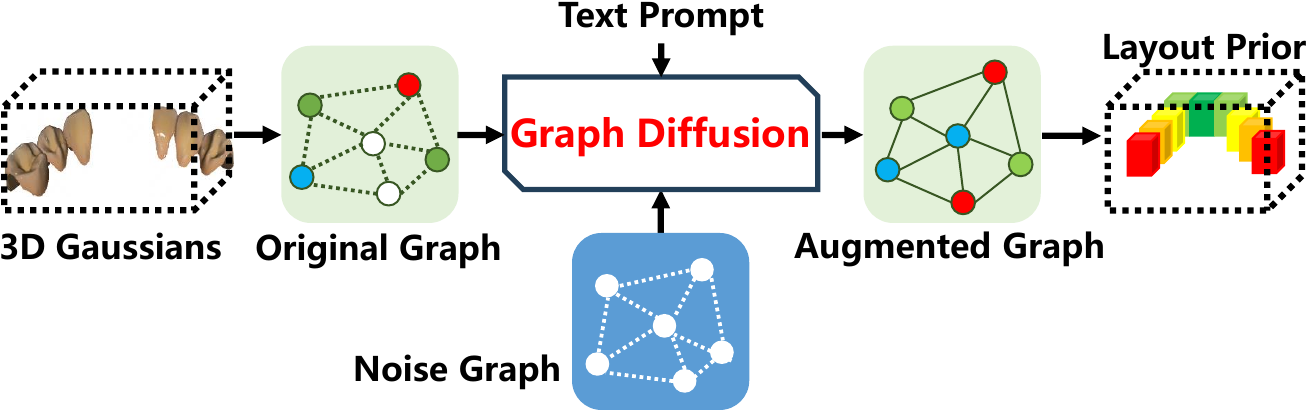}}
\caption{\textbf{Illustration of approaches using the pairwise relations and the proposed approach.} 
(a) GALA3D uses only the text description to optimize the layout (teeth positions), receiving limited results. 
(b) Our DM-CFO constructs a graph to represent the jaw with multiple missing teeth, then the target graph is incrementally restored during the denoising process through a graph diffusion model.
}\label{figchallenges}
\end{figure}

Although contemporary text-to-3D models~\cite{poole2023dreamfusion} and image-to-3D models~\cite{ge2025compgs} demonstrate the capability to generate individual 3D teeth, they face significant challenges in the compositional generation of multiple 3D teeth~\cite{chen2024comboverse,yan2024frankenstein}, where multiple missing 3D teeth in a jaw are simulated using contextual information, such as geometric knowledge of neighboring teeth.
This difficulty arises from the need to optimize both layouts and shapes of missing teeth.
Large language models (LLMs), such as GPT-3.5, have been utilized to explicitly generate scene graphs~\cite{zhou2024gala3d,li2024discene}, which facilitate the supervision of shape generation for each individual instance.
An example is GALA3D~\cite{zhou2024gala3d}, illustrated in Fig.~\ref{figchallenges}(a).
However, this exploration is restricted to pairwise relationships, which inform the 3D synthesis but result in geometric inconsistencies among objects due to the lack of higher-order information.
Additionally, collision conflicts are frequently overlooked in 3D Gaussian–based compositional generation, as objects may intersect due to the absence of explicit geometric information regarding their surfaces.
DreamScape~\cite{yuan2024dreamscape} introduces a collision-loss mechanism that calculates the aggregate distance between closely located points of two objects, using a predetermined threshold.
Nevertheless, this approach is insufficient for instances characterized by varying scales.

In recent years, denoising diffusion models~\cite{ho2020denoising} have exhibited exceptional generative capabilities in the domain of graph generation~\cite{kong2023autoregressive}.
These models offer several advantages, including stable training processes and the ability to generalize across various graph structures.
Consequently, graph diffusion–based models~\cite{kong2023autoregressive} can be utilized effectively in the design of compositional tooth models by optimizing the arrangement of missing teeth within the jaw.
Moreover, since a tooth can be approximated as a cylindrical shape, the spatial relationships between 3D Gaussians corresponding to different teeth can be leveraged to impose penalties for collision conflicts.

In this study, we propose an approach named DM-CFO, where the jaw configuration, including absent teeth, is generated through a graph diffusion model, in which a target graph is incrementally restored during the denoising process by integrating both textual and graphical constraints, as illustrated in Fig.~\ref{figchallenges}(b).
Following this, we alternately update the Gaussian parameters of each layout-guided tooth, as well as those of the entire jaw, using score distillation sampling (SDS)~\cite{poole2023dreamfusion}.
A regularization term, based on the distances between the 3D Gaussian representations of neighboring teeth and the anchor tooth, is introduced to mitigate tooth intersections.

The principal contributions of this paper are summarized as follows.
\begin{itemize}
\item A novel framework utilizing 3D Gaussian splatting is proposed for the automatic generation of multiple missing teeth. This framework operates by alternately updating the Gaussian parameters associated with both the overall scene and the individual instances.
\item The configuration of teeth within a jaw, which may include the occurrence of several missing teeth, is produced utilizing a graph diffusion model. In this model, a target graph is systematically reconstructed during the denoising phase employing both textual and graphical constraints.
\item A collision loss based on 3D Gaussians is proposed to penalize intersections between teeth, thereby improving the geometric quality of the designed tooth model.
\end{itemize}

The experimental results obtained from Shining3D~\cite{wang2022tooth}, Aoralscan3~\cite{tian2023revised}, and DeepBlue~\cite{tian2024rgb} tooth design datasets demonstrate that the methodology presented in this article is competitive with current state-of-the-art (SOTA) techniques in the design of compositional tooth models.

The remainder of this paper is organized as follows:
Section~\ref{relatedWork} reviews studies on 3D generation and editing, 
Section~\ref{ourApproach} introduces the proposed approach,
Section~\ref{results} presents the experimental results,
and Section~\ref{conclusion} concludes the article.

\section{Related Work}    \label{relatedWork}
In this section, we provide a succinct review of the literature on 3D generation and editing, with particular emphasis on diffusion models and 3D Gaussian splatting representations.

\subsection{3D Tooth Generation}
The field of 3D tooth generation is driven by the integration of deep learning techniques that automate and personalize restorative design.
Traditional methods employ the encoder-decoder architecture to reconstruct an integrated point cloud when a partial point cloud is provided. 
For example, VF-Net~\cite{ye2023variational} is a fully probabilistic point cloud model closely resembling variational autoencoders to replace the Chamfer distance and enable working with probability densities.
Recently, research efforts are increasingly focused on transformer-based architectures and multimodal data fusion.
For example, TranSDFNet~\cite{shen2023transdfnet} introduces a voxel-based truncated signed distance field (TSDF) to improve smooth reconstruction.
The point-to-mesh completion network~\cite{hosseinimanesh2025personalized} generates watertight meshes directly from partial scans, leveraging implicit neural representations for improved marginal fit.
SSEN~\cite{shi2025self} learns directly from unlabeled dental data by identifying multimodal features and topological relationships.
VBCD~\cite{wei2025vbcd} employs a coarse-to-fine architecture and loss of curvature and margin line to reconstruct dental crowns.

Generative adversarial networks (GANs)~\cite{chafi2025exploring} are another important architecture for 3D tooth generation, adding a discriminator to evaluate the effectiveness of distribution simulation.
A two-stage GAN framework~\cite{roh2024two} divides the generation task into segmentation and depth estimation stages to improve morphological consistency, while a three-stage architecture~\cite{wu2025automatic} adds an additional image inpainting stage to handle challenging cases.
MVDC~\cite{yang2025mvdc} synthesizes occlusal, buccal, and lingual depth maps to reconstruct crown geometries with high fidelity, emphasizing holistic shape inference.

Diffusion models have recently been used for the synthesis of photorealistic textures, which improve aesthetic outcomes by generating realistic enamel surfaces~\cite{saleh2024feasibility,wang2025diff}.
However, multiple-tooth generation is still underexplored, since both tooth layout and tooth surface information are required in the inference stage, which increases the complexity of the generation task.

\subsection{3D Gaussian Splatting-based Generation and Editing}
3D Gaussian Splatting (3DGS) utilizes millions of Gaussian ellipsoidal point clouds to accurately represent objects or scenes and facilitate view rendering through rasterization.
To address computational initialization, LGM~\cite{tang2024lgm} proposes a regression network that directly initializes Gaussian parameters from multiview images end-to-end.
Moreover, 3DGS is limited in surface modeling due to the intrinsic properties of its representations, particularly in scenarios involving multiple instances within the target scene.
To mitigate this issue, a mesh is extracted from 3DGS using a local density query~\cite{tang2024dreamgaussian}.

In the context of image editing, the rendered image and textual embeddings are generally aligned to accurately identify the target within the rendered image~\cite{wang2024gaussianeditor}.
Certain methodologies~\cite{zhuang2024tip} attribute inaccuracies in controlling the specified appearance and location to the inherent limitations of text descriptions.
These approaches accommodate both text and image prompts simultaneously to delineate the editing region.
Furthermore, Gaussian semantic features~\cite{chen2024gaussianeditor,ye2024gaussian} are utilized to identify or track the edited region from multiple viewpoints.

Collision conflicts are frequently overlooked in 3D Gaussian–based compositional tooth generation, as teeth may intersect due to the absence of explicit geometric information regarding their surfaces.
DreamScape~\cite{yuan2024dreamscape} introduces a collision-loss mechanism that calculates the aggregate distance between closely situated points of two objects using a predetermined threshold.
Nevertheless, this approach is insufficient for teeth characterized by varying scales.

\subsection{Diffusion Model-based Generation and Editing}
Significant advances have been made in 3D generation and editing, largely attributable to the rapid development of diffusion models~\cite{rombach2022high}.
Direct 3D editing utilizes semantic 3D representations, such as neural shape representations, to manipulate the appearance, shape, or existence of target objects.
The optimization of both the entire scene and its zoomed-in sections is performed jointly to address the multi-face problem and improve detail~\cite{cao2024dreamavatar}.

In scenarios where limited 3D scene data are available, various strategies~\cite{poole2023dreamfusion} have been developed to improve 3D representations by leveraging prior information extracted from 2D diffusion models.
To address the issue of multiview inconsistency, multiview diffusion models~\cite{gao2024cat3d,chen2024v3d,han2024vfusion3d} are employed to generate coherent images from novel target viewpoints.

To generate objects characterized by intricate textures and complex geometries, it is essential to integrate global features with local features~\cite{yang2024magic}.
This integration provides precise guidance for the SDS, which aligns effectively with the input data.
However, extending this approach to the generation of multiple missing teeth presents challenges, as the local patterns of distinct teeth are independent of each other, and neighboring teeth may encounter collision conflicts due to the lack of contextual information from their surroundings.

\begin{figure*}[htbp]
\centering
\includegraphics[width=0.9\linewidth]{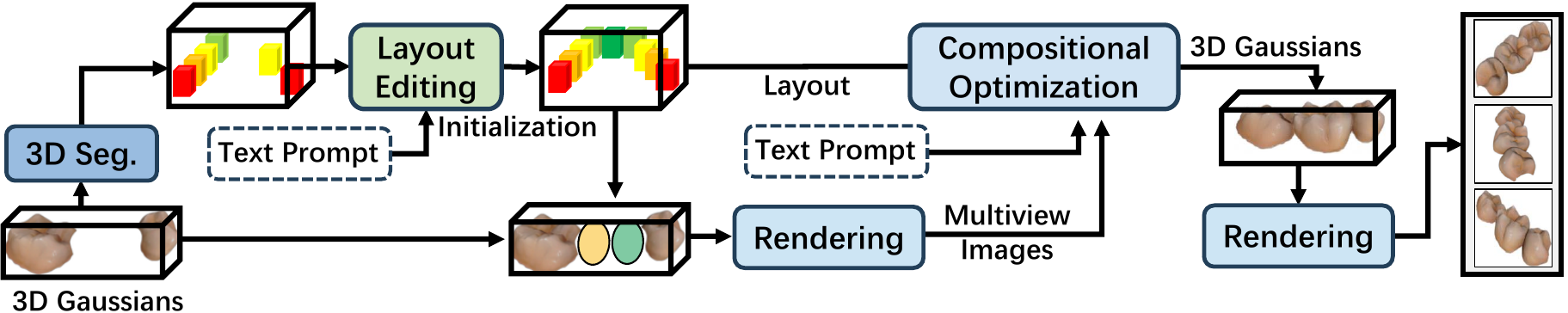} \\
\caption{\textbf{An illustration of the proposed DM-CFO.}
Given 3D Gaussian representing jaw with missing teeth, the graph diffusion generates layout of missing teeth by progressively denoising graph with both text and graph constraints.
Then, the Gaussian parameters of each layout-guided tooth and the whole jaw are alternately updated using SDS. 
A regularization term based on 3D Gaussians of neighboring teeth is explored to penalize the tooth intersection.
}\label{figFramework}
\end{figure*}

\subsection{Compositional 3D Generation}
Compositional 3D generation, which entails the synthesis of scenes comprising multiple instances, poses considerable challenges within 3D contexts.
Certain 3D reconstruction methodologies, such as Comboverse~\cite{chen2024comboverse} and REPARO~\cite{han2025reparo}, independently segment, complete, and generate multiple instances, subsequently optimizing their spatial relationships by aligning them with reference images.
An alternative approach~\cite{ge2025compgs} emphasizes compositional text-to-3D generation, in which the reference image is generated by a text-to-image model.
Although Deep Prior Assembly (DPA)~\cite{zhou2024zero} and the Divide-and-Conquer (DAC) strategy~\cite{dogaru2024generalizable} utilize depth priors to aid in scene assembly, they face challenges in compositional 3D generation~\cite{chen2024comboverse,yan2024frankenstein}, as both the layout of the scene and the shapes of the instances require optimization.

Layout-your-3D~\cite{zhou2024layout} adopts 2D layouts to enhance space-aware SDS, ensuring precise control over the generation process.
Nevertheless, reliance on 2D guidance often presents challenges in accurately composing multiple objects with diverse attributes and interrelationships into a cohesive scene.
To address the differentiation of various attributes within implicit 2D diffusion priors, GALA3D~\cite{zhou2024gala3d} generates a coarse 3D layout prior and subsequently learns a layout-guided Gaussian representation.
Within this framework, MVDream~\cite{shi2024mvdream} functions as the instance-level diffusion prior, while ControlNet~\cite{zhang2023adding} operates as the scene-level diffusion prior.
SceneWiz3D~\cite{zhang2024towards} employs particle swarm optimization (PSO) to refine the layout of the scene.
In addition, large language models, such as GPT-3.5, are employed to explicitly generate scene graphs~\cite{gao2024graphdreamer,zhou2024gala3d,li2024discene} that supervise the shape generation of each instance.
However, the current approach relies solely on adjacency relations to guide 3D synthesis, which results in geometric inconsistencies among instances due to the lack of higher-order information.

To enhance the understanding of relationships among instances, MIDI~\cite{huang2024midi} introduces a multi-instance attention mechanism that effectively captures complex inter-object interactions.
Simultaneously, ComboVerse~\cite{chen2024comboverse} modifies the attention map of position tokens, which represent spatial relationships for score distillation.
However, the issue of collision conflicts remains unaddressed in 3D Gaussian–based compositional generation, where objects may intersect due to the absence of explicit geometric information regarding their surfaces.
PhyCAGE~\cite{yan2024phycage} introduces an innovative refinement of SDS by integrating physics-based simulation. Rather than relying solely on traditional gradient updates, the method repurposes the SDS loss gradient as the initial velocity in a dynamic physical simulation.
DreamScape~\cite{yuan2024dreamscape} proposes a collision-loss mechanism that calculates the sum of distances between points that are in close proximity across two objects, based on a predetermined threshold.
However, this approach proves inadequate for instances with varying scales, particularly in scenarios that require stringent collision rejection.

\section{The Proposed Approach}  \label{ourApproach}
3D Gaussian splatting, diffusion models, and score distillation sampling collectively establish a foundational framework for 3D editing.
In the context of a jaw model that exhibits multiple missing teeth, we construct and optimize the arrangement of all teeth using a graph diffusion model.
We propose a dual-level optimization approach aimed at achieving instance-level realism and global consistency, thereby enhancing the fidelity of the synthesized teeth while mitigating collision conflicts.
The details of our DM-CFO are illustrated in Fig.~\ref{figFramework}.

\subsection{Preliminaries} \label{Preliminaries}
\textbf{3D Gaussian Splatting}.
3DGS~\cite{kerbl20233d} represents an explicit radiance field that employs anisotropic 3D Gaussians for scene representation, thereby enabling high-quality, real-time, and high-resolution rendering. 
To enhance rendering efficiency, 3DGS incorporates a tile-based rasterizer that segments the image into tiles, which are subsequently filtered and sorted according to the projected 3D Gaussians. The color $C$ of each pixel within the patch is defined as follows:
\begin{equation} \label{eq1}
C = \sum_{i \in \mathcal{N}} c_i \alpha_i \prod^{i-1}_{j=1}(1-\alpha_j),
\end{equation}
where $c_i$ is derived from spherical harmonics, which characterize the color of a point among the $\mathcal{N}$ ordered points that intersect with the pixel, and $\alpha$ is the opacity of the point, which is scaled by a 2D Gaussian.

\textbf{Diffusion Model}.
Diffusion models are a category of generative models that systematically denoise samples $\mathbf{I}$ initially drawn from a Gaussian distribution, with the ultimate goal of aligning these samples with the real data distribution $p(\mathbf{I})$. 
These models are based on two primary processes. The forward process utilizes Gaussian noise scheduling to perturb the data, which is represented as $\mathbf{I} \sim p(\mathbf{I})$. 
A reverse process incrementally removes this noise and reintroduces structure into an intermediate latent variable, expressed as $\mathbf{I}_{\eta} = \alpha_{\eta} \mathbf{I} + \sigma_{\eta} \epsilon$, where $\alpha_{\eta}$ and $\sigma_{\eta}$ are noise schedules at timestep $\eta$, and $\epsilon$ is Gaussian noise. The reverse process is typically parameterized by a conditional neural network $\mathbf{\epsilon}_{\mathbf{\phi}}$ that is trained to predict the noise $\epsilon$ using a simplified objective function.


\textbf{Score Distillation Sampling}.
SDS utilizes a pre-trained DM to improve the optimization of a differentiable and parametric image rendering function $g(\mathbf{\theta}, \mathbf{\pi})$, where $\mathbf{\theta}$ is a 3D representation, and $\mathbf{\pi}$ denotes the camera pose from which the image $\mathbf{I}$ is produced. Specifically, the parameters $\mathbf{\theta}$ are updated by employing the gradient:
\begin{equation}   \label{eq2}
\triangledown_{\mathbf{\phi}} L_{SDS}(\mathbf{\phi}, \mathbf{\theta}) = \mathbb{E}_{\epsilon \sim \mathcal{N}(\mathbf{0}, \mathbf{I}), \eta \sim T}[\omega(\eta) (\hat{\mathbf{\epsilon}}_{\mathbf{\phi}}(\mathbf{z}_{\eta}, \eta, \mathbf{c}) - \mathbf{\epsilon}) \frac{\partial \mathbf{z}_{\eta}}{\partial \mathbf{\theta}}],
\end{equation}
where $\mathbf{z}_{\eta} = E(g(\mathbf{\theta}, \mathbf{\pi}))$ represents the latent representation obtained by the encoder $E$.

\subsection{Layout Editing Based on Graph Diffusion} \label{LayoutEditing}
Graph models are employed to represent spatial relationships among instances with the assistance of LLMs, they encounter difficulties in addressing complex intrinsic dependencies. 
Unlike image data, the nodes within these models do not adhere to the assumption of being independent and identically distributed (i.i.d.). The complexity inherent in the graph structure presents significant obstacles in the generation of the desired graphs.

\begin{figure}[htbp]
\centering 
\includegraphics[width=0.99\linewidth]{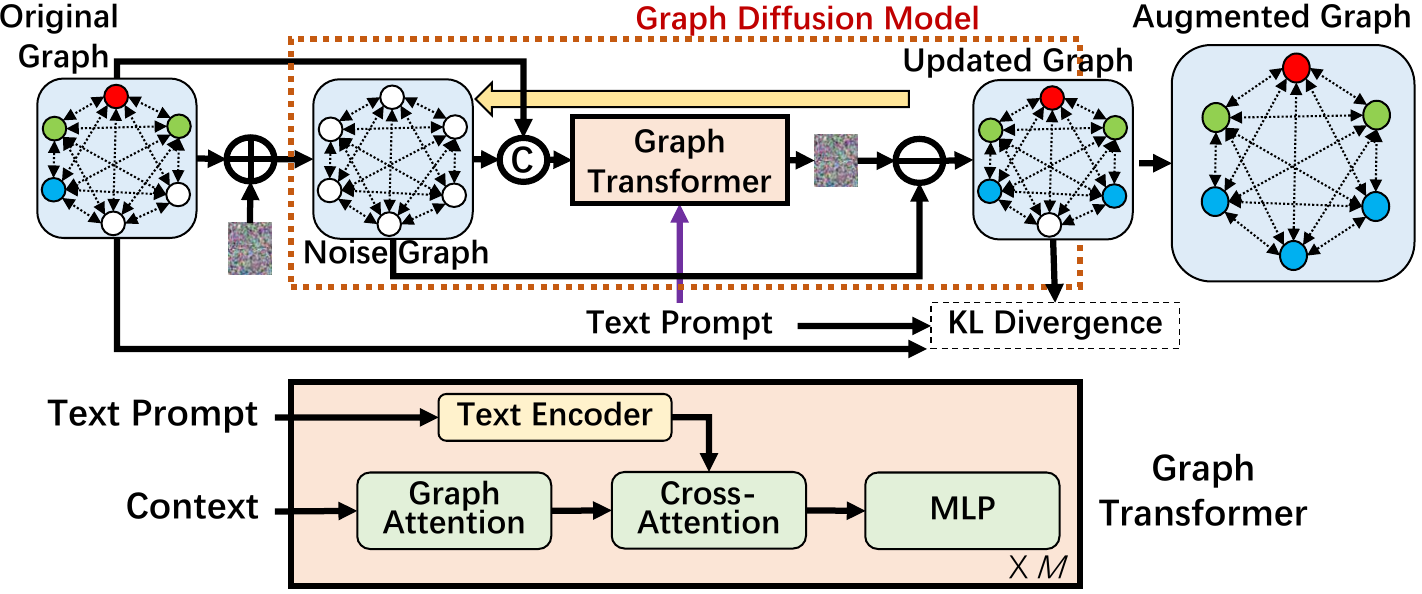} \\
\caption{\textbf{An illustration of the layout editing module.}
Given the original jaw graph, noise is progressively added in the forward diffusion process, and then graph is progressively updated in the reverse denoising process to obtain the augmented jaw graph. The yellow arrow represents the iteration step.}\label{figLayoutEditing}
\end{figure}

Drawing upon recent advances in language-guided 3D scene synthesis, we transform scenes into semantic graphs and utilize a conditional diffusion model to learn the conditional distribution of the target scene graph, which is illustrated in Fig.~\ref{figLayoutEditing}.

In this study, a 3D jaw model characterized by multiple missing teeth is used to facilitate the segmentation of each individual tooth using a commercially available approach~\cite{tian20223d}. 
An original jaw graph $\mathcal{G}_s=(\mathbf{V}_s, \mathbf{E}_s)$ is constructed to represent the jaw layout $\mathbf{L}_s$, where each node $\mathbf{v}_i \in \mathbf{V}_s$ represents a specific tooth. 
Each node encompasses a discrete category \(c_i\), a tooth layout $\mathbf{L}_i$, and semantic features $\mathbf{f}_i$ derived from the segmentation process.
The tooth layout $\mathbf{L}_i = \{ x_i, y_i, z_i, h_i, w_i, l_i, k_i, r_i \}$, where $(x_i, y_i, z_i)$ represent the spatial coordinates of the $i$-th tooth; $(h_i, w_i, l_i)$ are the length, width, and height of the tooth boundary, $k_i$ and $r_i$ are the rotation angles of the dental and buccal sides.
The discrete category $c_i$ of missing teeth is based on established dental knowledge. 
In contrast, the corresponding tooth layout $\mathbf{L}_i$ and semantic features $\mathbf{f}_i$ are initially set to `$\varnothing$' and are progressively refined throughout the optimization process.
Each edge $\mathbf{e}_{ij} \in \mathbf{E}_s$ corresponds to a unique type of relationship, which can be specifically classified into the following categories: [`Neighbor', `Symmetry', `Arch'].
Therefore, $\mathcal{G}_s = (\mathbf{C}_s, \mathbf{L}_s, \mathbf{F}_s, \mathbf{E}_s)$, where $\mathbf{C}_s$, $\mathbf{L}_s$, $\mathbf{F}_s$, and $\mathbf{E}_s$ encapsulate the aggregated classifications, layouts, and features of all teeth, as well as the interrelationships among teeth within a given jaw.

Subsequently, the original jaw graph denoted $\mathcal{G}_s$ is optimized to produce the augmented jaw graph $\mathcal{G}_t = (\mathbf{C}_t, \mathbf{L}_t, \mathbf{F}_t, \mathbf{E}_t)$ using a jaw text prompt $\mathbf{y}_s$ via a discrete graph diffusion model $\epsilon_g$, where $\mathbf{C}_t$, $\mathbf{L}_t$, $\mathbf{F}_t$, and $\mathbf{E}_t$ represent the assembled classes, layouts, features of all teeth, and the interrelationships between the teeth within the jaw, respectively.
The objective of this module is to examine the conditional distribution $q(\mathcal{G}_t | \mathcal{G}_s, \mathbf{y}_s)$.
In the diffusion phase, Gaussian noise is incrementally introduced to the augmented jaw graph $\mathcal{G}_t$ to obtain $\mathcal{G}^{\eta}_t$ at the timestep $\eta$.
In the denoising phase, the graph diffusion model $\epsilon_g$ systematically reconstructs the image $\mathcal{G}^0_t$ using the control signal $\mathcal{G}_s$ and the vector $\mathbf{y}_s$.
Each element of the source scene graphs is concatenated with the noisy target scene graphs to provide contextual information, and then a graph Transformer with a frozen text encoder iteratively removes noise and updates graph configurations.
The graph Transformer consists of a stack of $M$ blocks, each comprising graph attention, cross-attention, and multilayer perceptron (MLP), where cross-attention layers are utilized to integrate linguistic features.
Assume that
\begin{equation} \label{eq10}
    L_{\eta-1} = D_{KL}[q(\mathcal{G}^{\eta-1}_t | \mathcal{G}^{\eta}_t, \mathcal{G}^0_t) || p_{\epsilon_g} (\mathcal{G}^{\eta-1}_t | \mathcal{G}^{\eta}_t, \mathcal{G}_s, \mathbf{y}_s)],
\end{equation}
where $D_{KL}$ indicates the KL divergence. Then, the variational lower bound (VLB) of the likelihood is derived as follows:
\begin{equation} \label{eq10}
    L_g = \mathbb{E}_{q(\mathcal{G}^0_t)} [\sum^T_{\eta=2} L_{\eta-1} - \mathbb{E}_{q(\mathcal{G}^1_t | \mathcal{G}^0_t)} [\log p_{\epsilon_g} (\mathcal{G}^0_t | \mathcal{G}^1_t, \mathcal{G}_s, \mathbf{y}_s)] ].
\end{equation}

During the inference phase, the layout $\mathbf{L}_s$ and the corresponding original scene graph $\mathcal{G}_s$ are generated based on the results of 3D segmentation. Subsequently, conditioned on the original scene graph $\mathcal{G}_s$ and the textual prompt $\mathbf{y}_s$, the graph diffusion model $\epsilon_g$ predicts the augmented scene graph $\mathcal{G}_t$.

The optimization details are presented in Algorithm~\ref{alg:algorithm1}. 
Our approach provides two notable advantages. 
First, the structural dependencies of complex graphs are effectively modeled and optimized through the noising and denoising phases of the diffusion model. 
Furthermore, despite the inherent discreteness of the graph structure, which complicates the computation of model gradients, continuous noise is integrated during the backpropagation training process for graph generation.

\begin{algorithm} \label{alg:algorithm1}
    \caption{Layout Optimization Based on Graph Diffusion}    
    \KwIn{Original jaw graph $\mathcal{G}_s = (\mathbf{V}_s, \mathbf{E}_s)$ with node attributes $(\mathbf{C}_s, \mathbf{L}_s, \mathbf{F}_s)$, text prompt $\mathbf{y}_s$, pre-trained graph diffusion model $\epsilon_g$;}
    \KwOut{Augmented jaw graph $\mathcal{G}_t = (\mathbf{C}_t, \mathbf{L}_t, \mathbf{F}_t, \mathbf{E}_t)$;}	
    \BlankLine
    \textbf{Initialization:} Target graph $\mathcal{G}_t^0 \leftarrow \mathcal{G}_s$;
    \BlankLine
    \textbf{Forward Diffusion Process:}\\
    \For{$\eta = 1$ \KwTo $T$}{
        Add Gaussian noise: $\mathcal{G}_t^\eta \leftarrow \alpha_\eta \mathcal{G}_t^{\eta-1} + \sigma_\eta \epsilon$;\\
    }
\BlankLine
\textbf{Reverse Denoising Process:}\\
 \For{$\eta = T$ \KwTo $1$}{
        Concatenate context: $\mathcal{G}_t^{\eta} \leftarrow [\mathcal{G}_s; \mathcal{G}_t^{\eta}]$;\\
        Apply graph transformer:
        $\hat{\epsilon}_g \leftarrow \epsilon_g(\mathcal{G}_t^\eta, \eta, \mathcal{G}_s, \mathbf{y}_s)$;\\
        Update graph:
        $\mathcal{G}_t^{\eta-1} \leftarrow \frac{1}{\alpha_\eta}(\mathcal{G}_t^\eta - \sigma_\eta \hat{\epsilon}_g) + \sigma_\eta \mathbf{z}$;\\
        KL-divergence: $L_\eta \leftarrow D_{KL} (q(\mathcal{G}_t^{\eta-1}|\mathcal{G}_t^\eta,\mathcal{G}_t^0) \| p_{\epsilon_g})$;
    }
    \BlankLine
    \textbf{Post-Processing:} Extract final layout $\mathbf{L}_t$ from $\mathcal{G}_t^0$.
\end{algorithm}

\subsection{Compositional Optimization} \label{CompositionalOptimization}
The generation of multiple teeth presents a significant challenge, as it necessitates the optimization of both layout and geometry. 
This process employs implicit diffusion priors to maintain consistency among the various teeth while simultaneously preventing collision conflicts with adjacent teeth.

\begin{figure}[htbp]
\centering 
\includegraphics[width=0.99\linewidth]{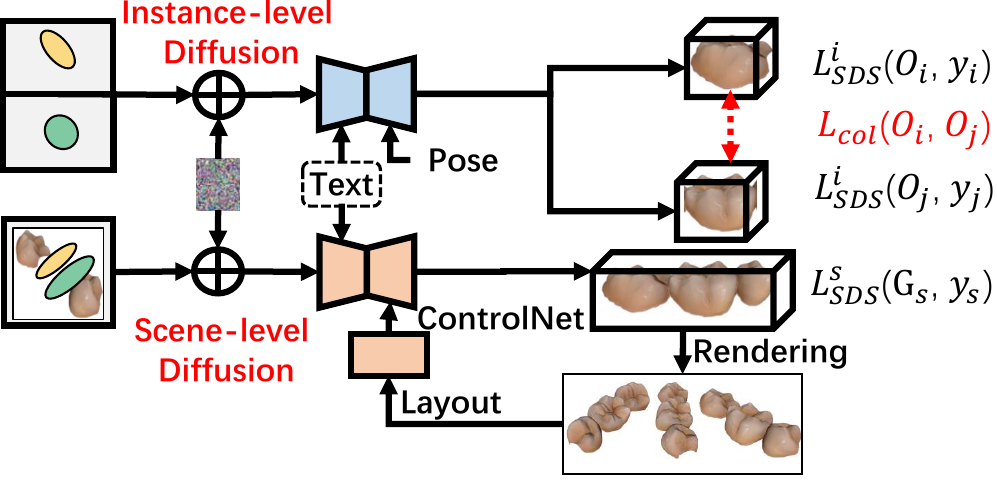} \\
\caption{\textbf{An illustration of the compositional optimization module.}
Instance-level diffusion and scene-level diffusion are jointly optimized.}\label{figCompositionalOptimization}
\end{figure}

Consequently, we propose a novel approach for synthesizing multiple teeth through the iterative optimization of the SDS loss, taking into account both scene and instance perspectives. 
This method employs multiview diffusion prior and incorporates heterogeneous constraints to enhance control, which is illustrated in Fig.~\ref{figCompositionalOptimization}.

Assume that the jaw geometry parameters $\mathbf{G}_s = \{ \mathbf{O}_i \}, i \in \{1,2, ..., N\}$ are composed of tooth geometry parameters $\mathbf{O}_i$ with the tooth index $i$ guided by the layout, and the tooth geometry parameters $\mathbf{O}_i = \{ \mathbf{L}_i, \mathbf{G}_i \}$ include the tooth layout $\mathbf{L}_i$ and the Gaussians of the tooth $\mathbf{G}_i$.
The tooth layout $\mathbf{L}_i$ was initially established based on the results of the graph diffusion.
The Gaussians of the tooth $\mathbf{G}_i$ are made up of anisotropic Gaussian functions. 
These functions are characterized by several parameters, including center $\mathbf{p}_i$, color $\mathbf{c}_i$, opacity $\alpha_i$, and covariance matrix $\mathbf{\Sigma}_i$.
The Gaussian splatting rendering~\cite{kerbl20233d} described in Eq.~\ref{eq1} is denoted by $g()$.
In each training iteration, the tooth Gaussians $\mathbf{G}_i$ are rendered to produce the tooth image $ \mathbf{I}^r_i = g(\mathbf{G}_i)$, while the jaw geometry parameters $\mathbf{G}_s$ are rendered to generate the jaw image $ \mathbf{I}^r_s = g(\mathbf{G}_s)$.
The gradient for the geometry parameters of the $i$-th tooth according to Eq.~\ref{eq2} can be expressed as follows:
\begin{equation} \label{eq8}
\triangledown_{\mathbf{O}_i} L^i_{SDS} = \mathbb{E}_{\epsilon, \eta}[\omega(\eta) (\epsilon_{\phi}(\mathbf{I}^r_i; \mathbf{y}_i, \mathbf{\pi}_i, \eta) - \epsilon) \frac{\partial \mathbf{I}^r_i}{\partial \mathbf{O}_i}],
\end{equation}
where $\epsilon$ represents the noise introduced, while $\eta$ denotes the time step. The function $\omega(\eta)$ serves as a weighting function. Additionally, $\epsilon_{\phi}$ refers to the denoising function employed in the image diffusion process. 
The variable $\mathbf{y}_i$ signifies the text prompt associated with the $i$-th tooth, and $\mathbf{\pi}_i$ indicates the extrinsic matrix of the camera.

Conditioned diffusion is used to improve the global scene by generating restored teeth while maintaining the original layout. 
In particular, ControlNet is fine-tuned to facilitate the rendering of layouts from multiple viewpoints as input, thereby producing 2D diffusion supervision that ensures consistency between layout and text. 
The gradient for the jaw geometry parameters can be articulated according to Eq.~\ref{eq2} as follows:
\begin{equation} \label{eq9}
\triangledown_{\mathbf{G}_s} L^s_{SDS} = \mathbb{E}_{\epsilon, \eta}[\omega(\eta) (\epsilon_{\phi}(\mathbf{I}^r_s; \mathbf{y}_s, \mathbf{\delta}_s, \eta) - \epsilon) \frac{\partial \mathbf{I}^r_s}{\partial \mathbf{G}_s}],
\end{equation}
where $\epsilon$ represents the noise introduced; $\eta$ denotes the time step; $\omega(\eta)$ serves as a weighting function; $\epsilon_{\phi}$ refers to the denoising function utilized in the diffusion process of 3DGS; $\mathbf{y}_s$ signifies the text prompt associated with the scene; and $\mathbf{\delta}_s$ represents the conditional input for the ControlNet, which is derived from rendering images based on the layouts.

However, the compositional loss is inadequate to address the significant collisions resulting from occlusion. 
Most existing methods rely on SDF to handle collision conflict~\cite{li2024focaldreamer}, and the conversion process between 3D Gaussians and SDF is intricate and not conducive to real-time applications. DreamScape~\cite{yuan2024dreamscape} presents a collision loss mechanism that utilizes 3D Gaussian representations. 
This methodology calculates the cumulative distances between points that are in close proximity across two objects, employing a predetermined threshold. Nevertheless, this approach proves inadequate for instances that exhibit variability in scale.

\begin{figure}[htbp]
\centering 
\includegraphics[width=0.8\linewidth]{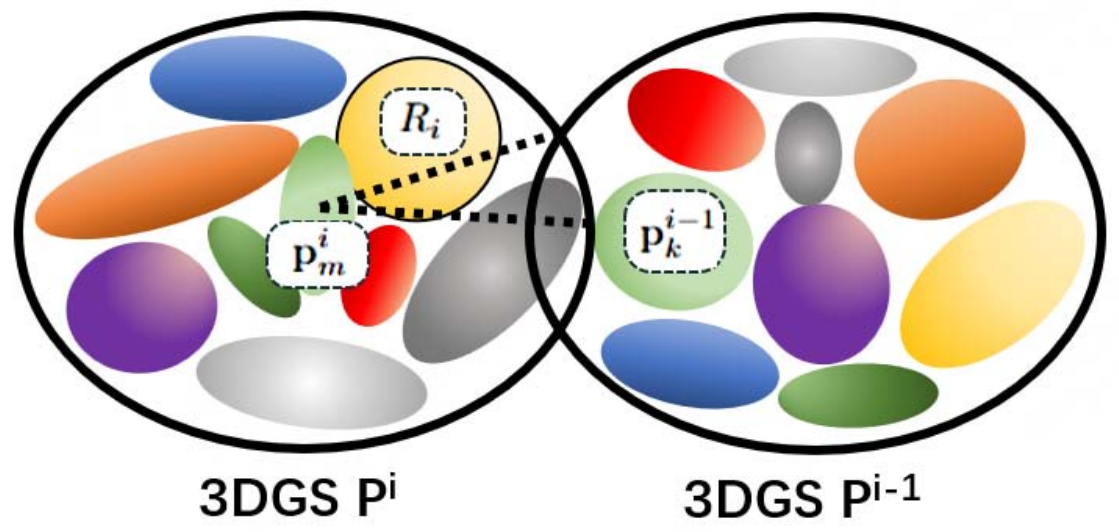} \\
\caption{\textbf{An illustration of collision loss.}
Our approach employs an intravariance $R_i$ rather than a fixed threshold to avoid collision conflict.}\label{figCollisionAvoidance}
\end{figure}

To address this challenge, we propose a collision loss mechanism grounded in a 3D Gaussian representation, which serves to distinguish between improperly overlapping instances. 
This methodology iteratively evaluates the distances from the 3D Gaussian representations of adjacent teeth to the anchor tooth, thereby imposing penalties for tooth intersections. 
An illustration of the proposed collision loss is presented in Fig.~\ref{figCollisionAvoidance}.

Assume that the Gaussians $\mathbf{P}^i = \{ \mathbf{p}^i_1, \mathbf{p}^i_2, ..., \mathbf{p}^i_{K_i} \}$ for the \textit{i}-th tooth are represented as a combination of Gaussian coordinates $\mathbf{p}^i_k$, where $k$ is the Gaussian index and $K_i$ is the number of Gaussians that comprise the \textit{i}-th tooth.
Initially, the mean coordinate of the \textit{i}-th tooth is determined as $\mathbf{p}^i_m$, and the intravariance $R_i = \frac{1}{K_i} \sum^{K_i}_{k=1} || \mathbf{p}^i_k - \mathbf{p}^i_m ||_2$ is calculated to evaluate the sparsity of the Gaussian distributions associated with the \textit{i}-th tooth.
Subsequently, the mean distances between the distributions of neighboring teeth are computed.
Specifically, the points associated with the \textit{i-1}-th and \textit{i+1}-th teeth are utilized to compute the distances to $\mathbf{p}^i_m$ and are then compared to the intra-variance $R_i$ to impose penalties for collisions occurring between adjacent teeth.
\begin{equation} \label{eq10}
\begin{split} 
     L^i_{col} = \sum_{k=1}^{K_{i-1}} \max(0, R_i - || \mathbf{p}^{i-1}_k - \mathbf{p}^i_m ||_2) +  \\ \sum_{k=1}^{K_{i+1}} \max(0, R_i - || \mathbf{p}^{i+1}_k - \mathbf{p}^i_m ||_2).
     \end{split}
\end{equation}
When two teeth are in conflict, the distance from the points within the intersection region to the center of the affected tooth is less than the intravariance of that tooth.

Given the tooth Gaussian loss $L^i_{SDS}$, the jaw Gaussian loss $L^s_{SDS}$, the collision regulation term $L_{col}$, the total loss is summarized as
\begin{equation} \label{eq11}
L^{total} = \lambda_1 \sum_{i=1}^{N} L^i_{SDS} + \lambda_2 L^s_{SDS} + \sum_{i=1}^{N} L^i_{col},
\end{equation}
The weights $\lambda_1$ and $\lambda_2$ balance the influence of various terms and are determined through a grid search methodology.

The intravariance $R_i$ is not a fixed hyperparameter but a property learned and optimized along with the Gaussian parameters (position, color, opacity, covariance) for each tooth. 
During SDS optimization, if teeth begin to intersect, the points in the overlapping region will cause the collision loss $L^i_{col}$ to increase. 
The gradient from this loss will push the Gaussians of both teeth apart. Consequently, the positions $\mathbf{p}^i_k$ are updated to minimize $L^i_{col}$, which inherently influences the calculated $R_i$ for the next iteration. 
This creates a feedback loop that continuously refines the tooth shape and spacing to avoid collisions.

The optimization details are encapsulated in Algorithm~\ref{alg:algorithm2}. Our methodology presents two main advantages. 
Firstly, the dual-level optimization facilitates the attainment of instance-level realism alongside global consistency, which enhances the fidelity of the synthesized dental structures. 
Secondly, it mitigates the issues of intersection and misalignment among the teeth, thereby addressing the potential spatial biases that may arise from the 3D Gaussian representations produced by LLMs and ensuring physical accuracy.

\begin{algorithm} \label{alg:algorithm2}
    \caption{Compositional Optimization for Scene and Instance}    
    \KwIn{Original jaw Gaussians $\mathbf{G}_s = \{\mathbf{O}_i\}$, target layout $\mathbf{L}_t$, text prompts $\{\mathbf{y}_s, \mathbf{y}_i\}$;}
    \KwOut{Optimized scene Gaussians $\mathbf{G}_s^{final}$ and instance Gaussians $\{\mathbf{G}_i^{final}\}$;}	
    \BlankLine
    Initialize teeth Gaussians $\{\mathbf{G}_i\}$ using layout $\mathbf{L}_t$\;
    \BlankLine
    \While{not converged}{
        \textbf{Scene-level Optimization:}\\
        \For{each camera view $\pi \in \Pi$}{
            Render scene image $\mathbf{I}_s^r = g(\mathbf{G}_s \cup \{\mathbf{G}_i\}, \mathbf{\pi})$\;
            Compute SDS loss $L_{SDS}^s$ via Eq.(\ref{eq9})\;
            Update $\mathbf{G}_s$ using $\nabla_{\mathbf{G}_s} L_{SDS}^s$\;
        }
        
        \textbf{Instance-level Optimization:}\\
        \For{each missing tooth $i \in L_t$}{
            \For{each camera view $\pi \in \Pi$}{
                Render instance image $\mathbf{I}_i^r = g(\mathbf{G}_i, \pi)$\;
                Compute SDS loss $L_{SDS}^i$ via Eq.(\ref{eq8})\;
                Compute collision loss $L_{col}^i$ via Eq.(\ref{eq10})\;
                Update $\mathbf{G}_i$ using $\nabla_{\mathbf{G}_i}(\lambda_1 L_{SDS}^i + L_{col}^i)$\;
            }
        }    
        Update layout $\mathbf{L}_t$ based on updated Gaussians.
    }
\end{algorithm}

\section{Results}    \label{results}
The effectiveness of the proposed methodology is evaluated and compared with its counterparts using the Shining3D~\cite{wang2022tooth}, Aoralscan3~\cite{tian2023revised}, and DeepBlue~\cite{tian2024rgb} tooth design datasets.

\subsection{Datasets and Evaluation Criteria}

The Shining3D tooth design dataset~\cite{wang2022tooth} consists of 1,416 meshes generated from 3D scans of dental plaster models, each obtained from a randomly selected patient in a dental hospital.
The dataset is divided into training, validation, and testing subsets, which contain 1,150, 133, and 133 samples, respectively.
A point cloud is extracted from each mesh, and an instance segmentation method~\cite{tian20223d} is used to classify and delineate the 3D region of each tooth.
Subsequently, the 3D regions of interest are cropped, including incisors, canines, and molars, while the remaining portions serve as conditions for simulation.
Each scene within the dataset comprises 60 images without visible teeth and 40 images that prominently feature them.
Masks representing the two-dimensional regions of the teeth of interest are generated using the approach described in~\cite{tian2020joint} and are made available for training and testing purposes.

\begin{figure}[htbp]
  \centering
  \subfigure[Age]{\includegraphics[width=0.31\linewidth]{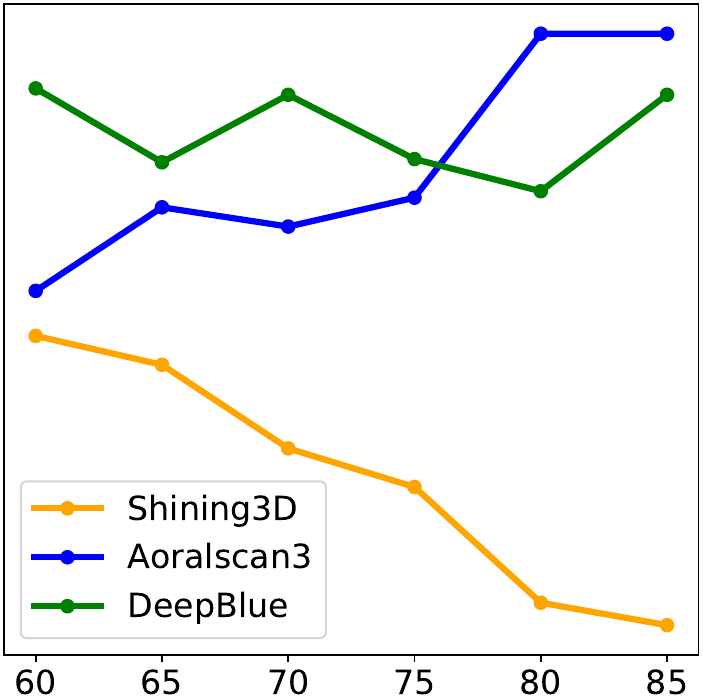}}
  \subfigure[Pathology]{\includegraphics[width=0.34\linewidth]{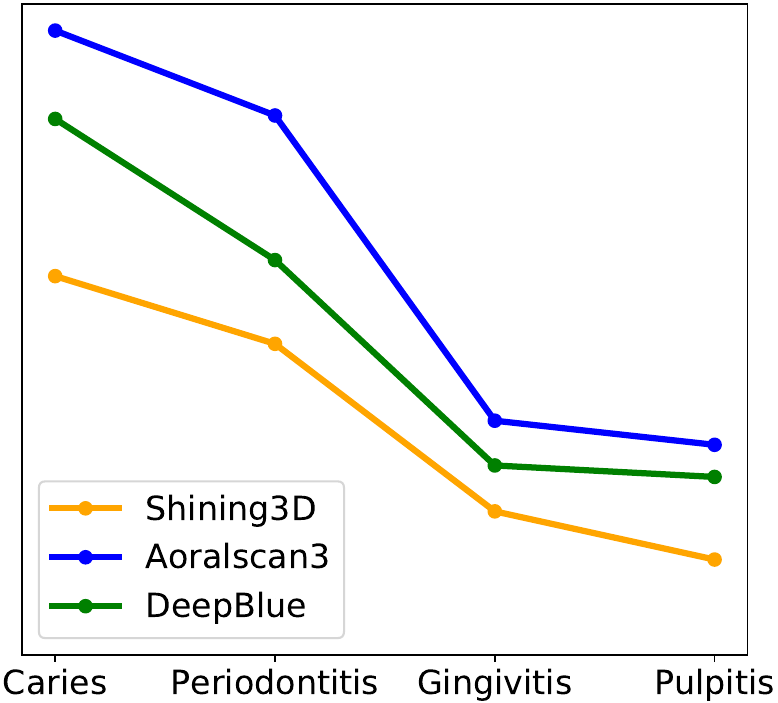}}
  \subfigure[Type]{\includegraphics[width=0.31\linewidth]{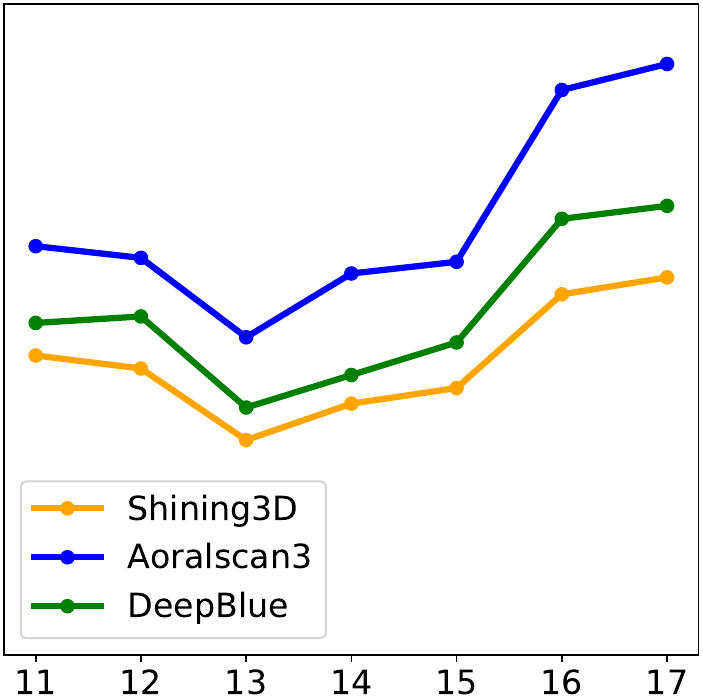}}
  \caption{\textbf{Data distributions in multiple datasets of (a) age, (b) pathology, and (c) types of missing teeth.}
}
  \label{fig_dataset}
\end{figure}

The Aoralscan3 dataset~\cite{tian2023revised} and the DeepBlue dataset~\cite{tian2024rgb} comprise 1,999 and 2,061 samples, respectively, with each sample derived from an anonymous patient.
While these datasets are primarily intended for the pose estimation of teeth, the 3D regions both with and without the teeth of interest can be utilized as paired data for the design of tooth models.
The construction of these datasets parallels that of the Shining3D tooth design dataset, in which each tooth is segmented using a distinct approach~\cite{zhao2026innovative}.
The Aoralscan3 dataset includes training, validation, and test sets containing 1,667, 156, and 176 samples, respectively, while the DeepBlue dataset comprises training, validation, and test sets containing 1,573, 244, and 244 samples, respectively.
Data distributions across multiple datasets, including age, pathology, and types of missing teeth, are illustrated in Fig.~\ref{fig_dataset}.

\begin{table*}
  \centering
  \caption{\textbf{Effectiveness comparison of the Shining3D, Aoralscan3, and DeepBlue dataset.}
  '$\downarrow$' means lower is better while '$\uparrow$' means upper is better.}
  \label{tabShining}
  \footnotesize 
  \begin{tabular}{lccccccccc}
\toprule
\multirow{2}{*}{Approach}  &\multicolumn{3}{c}{Shining3D}  &\multicolumn{3}{c}{Aoralscan3} &\multicolumn{3}{c}{DeepBlue} \\
\cmidrule(lr){2-4} \cmidrule(lr){5-7} \cmidrule(lr){8-10}
&FID$\downarrow$ &LPIPS$\downarrow$ &PSNR$\uparrow$   
&FID$\downarrow$ &LPIPS$\downarrow$ &PSNR$\uparrow$   
&FID$\downarrow$ &LPIPS$\downarrow$ &PSNR$\uparrow$ \\
\midrule
DGE~\cite{chen2024dge}           &223.15 &0.70 &12.38 &234.61 &0.75 &10.42 &228.73 &0.72 &11.11 \\
VcEdit~\cite{wang2024view}       &221.44 &0.69 &12.84 &233.67 &0.74 &11.02 &227.05 &0.71 &11.85 \\
Gaussctrl~\cite{wu2024gaussctrl} &220.90 &0.69 &13.09 &233.22 &0.74 &11.15 &226.10 &0.70 &12.20 \\
CAT3D~\cite{gao2024cat3d}        &218.50 &0.67 &14.45 &231.12 &0.73 &11.57 &224.24 &0.69 &13.08 \\
\midrule
CompGS~\cite{ge2025compgs}       &208.82 &0.65 &16.23 &216.75 &0.69 &13.10 &212.45 &0.67 &14.83 \\
Frankenstein~\cite{yan2024frankenstein} &205.43 &0.64 &17.06 &214.88 &0.68 &13.67 &208.57 &0.66 &15.39 \\
ComboVerse~\cite{chen2024comboverse}&202.43 &0.63 &17.57 &210.45 &0.67 &14.58 &206.52 &0.65 &15.91 \\
\midrule
DIScene~\cite{li2024discene}        &200.61 &0.62 &18.04 &209.72 &0.66 &15.61 &204.69 &0.64 &16.77 \\
DreamScape~\cite{yuan2024dreamscape}&198.83 &0.61 &18.87 &208.49 &0.65 &16.30 &203.34 &0.63 &17.60 \\
SceneWiz3D~\cite{zhang2024towards}  &198.59 &0.61 &18.90 &208.40 &0.65 &16.35 &203.40 &0.63 &17.68 \\
GALA3D~\cite{zhou2024gala3d}        &196.62 &0.60 &19.24 &206.44 &0.64 &17.14 &201.55 &0.62 &18.25 \\
MIDI~\cite{huang2024midi}           &195.71 &0.59 &20.39 &205.83 &0.63 &17.66 &200.77 &0.61 &18.95 \\
\textbf{Ours} &\cellcolor{blue!10}\textbf{193.29} &\cellcolor{blue!10}\textbf{0.57} &\cellcolor{blue!10}\textbf{22.55} 
&\cellcolor{blue!10}\textbf{203.56} &\cellcolor{blue!10}\textbf{0.61} &\cellcolor{blue!10}\textbf{19.02} 
&\cellcolor{blue!10}\textbf{198.41} &\cellcolor{blue!10}\textbf{0.59} &\cellcolor{blue!10}\textbf{20.74} \\
\bottomrule
\end{tabular}
\end{table*}

To assess the quality of the generated teeth, we compare the generated multiview images with the corresponding rendered views from other approaches by calculating the Peak Signal-to-Noise Ratio (PSNR), the Fréchet Inception Distance (FID), and the average Learned Perceptual Image Patch Similarity (LPIPS).
To quantitatively evaluate the results, we use the Chamfer Distance (CD) in millimeters (mm) and the F-Score to measure the similarity between the predicted teeth and the ground truth.
Moreover, to evaluate the effect of collision avoidance, we adopt the penetration distance (PD) of nearby teeth in millimeters (mm) as an additional evaluation metric.



\subsection{Implementation Details}
A workstation featuring an Intel i9-9980X 3.0 GHz CPU, 128 GB of RAM, and four NVIDIA RTX 4090D GPUs is used for performance evaluation.

In the layout editing phase, we use a 5-layer, 8-head graph Transformer with 512 attention dimensions and a dropout rate of 0.1.
We optimize the layout through 400 iterations. 

In the compositional optimization phase, the instance text prompt specifies the particular category of the generated tooth, while the scene text prompt consists of a combination of instance text prompts.
MVDream~\cite{shi2024mvdream} operates as a multiview diffusion model, utilizing a guidance scale of 50.
The ControlNet guidance scale is set to 100 to improve scene optimization and reduce the timestep during 3DGS. The learning rates for opacity and position are $5\times10^{-2}$ and $1.6\times10^{-4}$, respectively.
The color representation of the 3D Gaussians is achieved through spherical harmonic coefficients, with the degree fixed at 0.
The initial learning rate is set to $5\times10^{-3}$ and then attenuated to $5\times10^{-4}$ after epoch 380.
The covariance of the 3D Gaussians is decomposed into scaling and rotation for optimization, employing learning rates of $5\times10^{-3}$ and $10^{-3}$, respectively.
The coefficients $\lambda_1 = 10.0$ and $\lambda_2 = 2.5$ are determined using a grid-search methodology.
The stopping criterion is defined such that the training loss does not vary by more than 500 over 10 consecutive epochs.

\subsection{Ablation Study}
Extensive ablation studies were conducted to elucidate the impact of several critical components on the performance of the proposed methodology.
All experiments described in this subsection were performed using the Shining3D tooth design dataset.

\begin{figure}[htbp]
  \centering
  \subfigure[Tooth Number]{\includegraphics[width=0.4\linewidth]{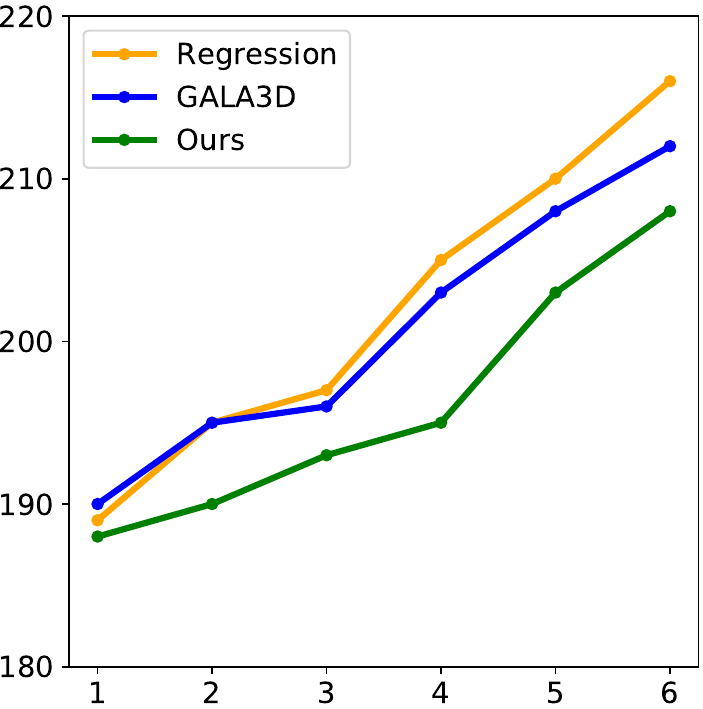}}
  \subfigure[Training Loss]{\includegraphics[width=0.4\linewidth]{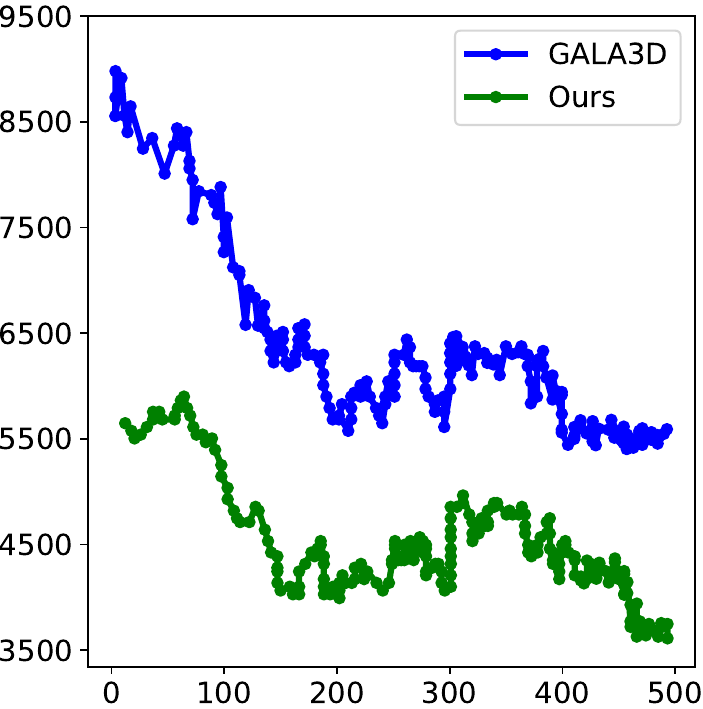}}
  \caption{\textbf{Illustration of variance on (a) tooth number and (b) training loss.}
}
  \label{fig_ab_class_num}
\end{figure}

\textbf{Tooth Number}.
The impact of the varying number of teeth produced by regression and by GALA3D, compared to our methodology, is illustrated in Fig.~\ref{fig_ab_class_num}(a).
The X-axis represents the class number, while the Y-axis represents the FID.
Layout errors and collisions increase significantly when using regression-based methods for multi-tooth generation.
The maximum number of simulated teeth is four, and the curvature of any additional teeth must take into account the dental arch; otherwise, the placement of the generated teeth may not satisfy the occlusal requirements.

\begin{figure}[htbp]
  \centering
  \includegraphics[width=0.9\linewidth]{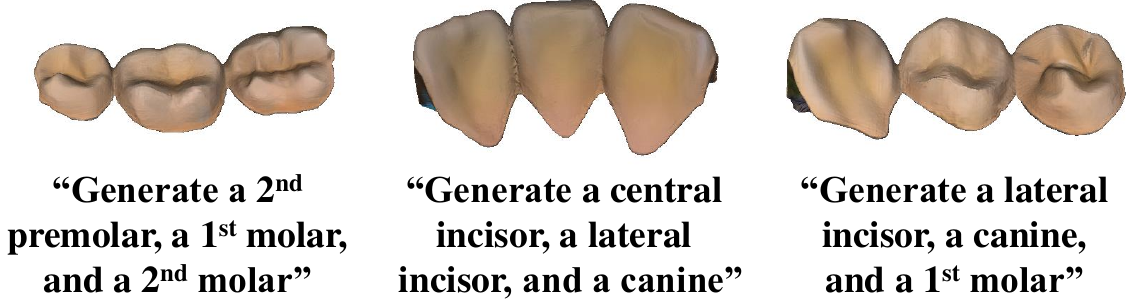}
  \caption{\textbf{Illustration of the impact of different group of tooth categories.} 
  Each column represents a specific group of tooth categories.}
  \label{fig_ab_ToothGroup}
\end{figure}

\textbf{Tooth Group}.
The impact of different categories of teeth on the formation of three distinct tooth types is depicted in Fig.~\ref{fig_ab_ToothGroup}, for example, a central incisor, a lateral incisor, and a canine.
Each column in the figure corresponds to a particular group of dental categories.
The findings indicate consistent robustness across the various categories of teeth.

\begin{figure*}[htbp]
\centering 
\includegraphics[width=0.90\linewidth]{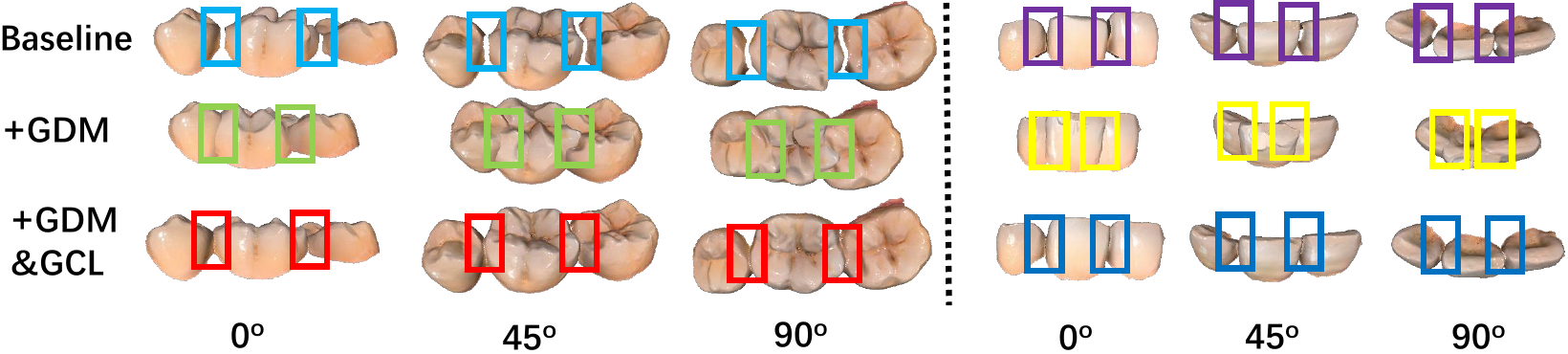}  
\caption{\textbf{Evaluation of the proposed modules on the Shining3D tooth dataset.}
Two samples are separated by the black dotted line. The baseline (GALA3D) refines layout without handling collisions. Integrating Graph Diffusion Model (GDM) improves tooth layout in a jaw, while adding Geometry Collision Loss (GCL) effectively resolves collisions, improving geometry quality.}\label{figEffectiveness}
\end{figure*}

\textbf{Training Loss}.
A comparison of the training loss between GALA3D and our proposed method is presented in Fig.~\ref{fig_ab_class_num}(b).
The X-axis denotes the epoch number, whereas the Y-axis represents the training loss.
Our methodology requires a greater number of epochs to converge, attributable to fluctuations in occlusion resulting from diverse viewpoints.
Nevertheless, the final loss achieved is lower than that of the alternative approach.

\textbf{Effectiveness}.
The efficacy of each component is detailed in Table~\ref{tab_effect} and illustrated in Fig.~\ref{figEffectiveness}.
The baseline utilizes the GALA3D architecture, which employs the layout interpreted by the LLM with subsequent refinement.
Note that no additional processing is implemented to address collision conflicts between instances.
As indicated in Table~\ref{tab_effect}, the graph diffusion model plays a crucial role in optimizing layouts by continuously adjusting them throughout the denoising process.
This methodology facilitates more intricately aligned interactions among instances while maintaining adherence to real-world constraints.
Furthermore, as shown in Fig.~\ref{figEffectiveness}, improvements in global scene optimization and geometric conflict resolution have resulted in the generation of 3D scenes that exhibit enhanced textures and scene coherence, effectively mitigating the occurrence of “over-constrained” boundaries.
\begin{table}[htbp]
  \centering
  \caption{\textbf{Effectiveness comparison of improved modules on the Shining3D dataset.}
  GDM represents the graph diffusion model, and GCL represents Gaussian collision loss in compositional optimization.}
  \label{tab_effect}
  \footnotesize 
  \begin{tabular}{ccccc}
\toprule
GDM         &GCL         &FID$\downarrow$ &LPIPS$\downarrow$ &PSNR$\uparrow$ \\
\midrule
            &             &196.62 &0.60 &19.24 \\
\checkmark  &             &194.25 &0.58 &21.85 \\
            &\checkmark   &194.49 &0.59 &20.71 \\
\checkmark  &\checkmark   &\cellcolor{blue!10}\textbf{193.29} &\cellcolor{blue!10}\textbf{0.57} &\cellcolor{blue!10}\textbf{22.55} \\
\bottomrule
\end{tabular}
\end{table}

\begin{table*}
  \centering
  \caption{\textbf{3D metric comparison of the Shining3D, Aoralscan3, and DeepBlue dataset.}
  '$\downarrow$' and '$\uparrow$' means lower and upper is better, respectively. '*' means the approach is implemented by ourselves.}
  \label{tab3D}
  \footnotesize 
  \begin{tabular}{lccccccccc}
\toprule
\multirow{2}{*}{Approach}  &\multicolumn{3}{c}{Shining3D}  &\multicolumn{3}{c}{Aoralscan3} &\multicolumn{3}{c}{DeepBlue} \\
\cmidrule(lr){2-4} \cmidrule(lr){5-7} \cmidrule(lr){8-10}
&CD$\downarrow$ &F-Score$\uparrow$ &PD$\downarrow$   
&CD$\downarrow$ &F-Score$\uparrow$ &PD$\downarrow$   
&CD$\downarrow$ &F-Score$\uparrow$ &PD$\downarrow$ \\
\midrule
TranSDFNet\cite{shen2023transdfnet}&0.33 &0.80 &0.16 &0.37 &0.77 &0.19 &0.36 &0.78 &0.18\\
Point-to-mesh\cite{hosseinimanesh2025personalized}*&0.28 &0.82 &0.14 &0.34 &0.79 &0.18 &0.30 &0.81 &0.16\\
SSEN~\cite{shi2025self}*       &0.25 &0.83 &0.14 &0.30 &0.81 &0.17 &0.27 &0.82 &0.16\\
VBCD~\cite{wei2025vbcd}       &0.24 &0.83 &0.12 &0.27 &0.81 &0.15 &0.26 &0.82 &0.14\\
\midrule
DPD~\cite{chafi2025exploring} &0.34 &0.79 &0.17 &0.38 &0.77 &0.21 &0.36 &0.78 &0.19\\
2Stage~\cite{roh2024two}*      &0.32 &0.80 &0.16 &0.37 &0.78 &0.20 &0.35 &0.79 &0.19\\
3Stage~\cite{wu2025automatic}* &0.31 &0.81 &0.15 &0.33 &0.79 &0.18 &0.32 &0.80 &0.17\\
MVDC~\cite{yang2025mvdc}      &0.28 &0.82 &0.14 &0.32 &0.80 &0.18 &0.30 &0.81 &0.16\\
\midrule
DM~\cite{saleh2024feasibility}&0.30 &0.81 &0.15 &0.32 &0.79 &0.17 &0.31 &0.80 &0.16\\
\textbf{Ours} &\cellcolor{blue!10}\textbf{0.22} &\cellcolor{blue!10}\textbf{0.86} &\cellcolor{blue!10}\textbf{0.07} 
&\cellcolor{blue!10}\textbf{0.26} &\cellcolor{blue!10}\textbf{0.84} &\cellcolor{blue!10}\textbf{0.10} 
&\cellcolor{blue!10}\textbf{0.24} &\cellcolor{blue!10}\textbf{0.85} &\cellcolor{blue!10}\textbf{0.09} \\
\bottomrule
\end{tabular}
\end{table*}

\subsection{Evaluation of the Tooth Design Datasets}
\textbf{Quantitative Comparison}.
A comparative analysis of various compositional 3D generation methodologies is conducted alongside our proposed method, utilizing identical parameters to ensure a fair evaluation, and the results are reported in Table~\ref{tabShining}.
The methodologies referenced in the literature~\cite{chen2024dge,wang2024view,wu2024gaussctrl,gao2024cat3d} generate multiple teeth by simulating a single tooth in each iteration.
This approach results in limited 3D consistency and incurs substantial computational time and resource costs.
Methods~\cite{ge2025compgs,yan2024frankenstein,chen2024comboverse} segment, complete, and generate multiple instances while neglecting layout information, which may lead to a deterioration in the quality of the generated output.
Methods~\cite{li2024discene,yuan2024dreamscape,zhang2024towards,zhou2024gala3d,huang2024midi} explore layouts using large language models or pairwise relations.
However, these methods may produce geometric inconsistencies among instances due to their inability to incorporate higher-order information.
In contrast, our approach demonstrates a significant improvement over all other methods in terms of FID and LPIPS.
For example, our method improves FID by 2.42$\%$ compared to MIDI~\cite{huang2024midi} on the Shining3D tooth design dataset, thereby underscoring its efficacy in compositional 3D generation.

The 3DGS are converted to mesh using Dreamgaussian~\cite{tang2024dreamgaussian}.
A comparative analysis of the quality of the generated mesh is conducted between different tooth generation methods, and the results are reported in Table~\ref{tab3D}.
Both encoder-decoder architecture~\cite{shen2023transdfnet,hosseinimanesh2025personalized,shi2025self,wei2025vbcd} and the GAN-based approaches~\cite{chafi2025exploring,roh2024two,wu2025automatic,yang2025mvdc} receive similar effectiveness.
Our approach improves CD by a margin of 6.0\%-8.0\% and PD by 7.0\%-8.0\% due to the layout prior, dual optimization of instance and scene, and collision avoidance regularization.

\begin{figure*}[htbp]
\centering 
\includegraphics[width=0.8\linewidth]{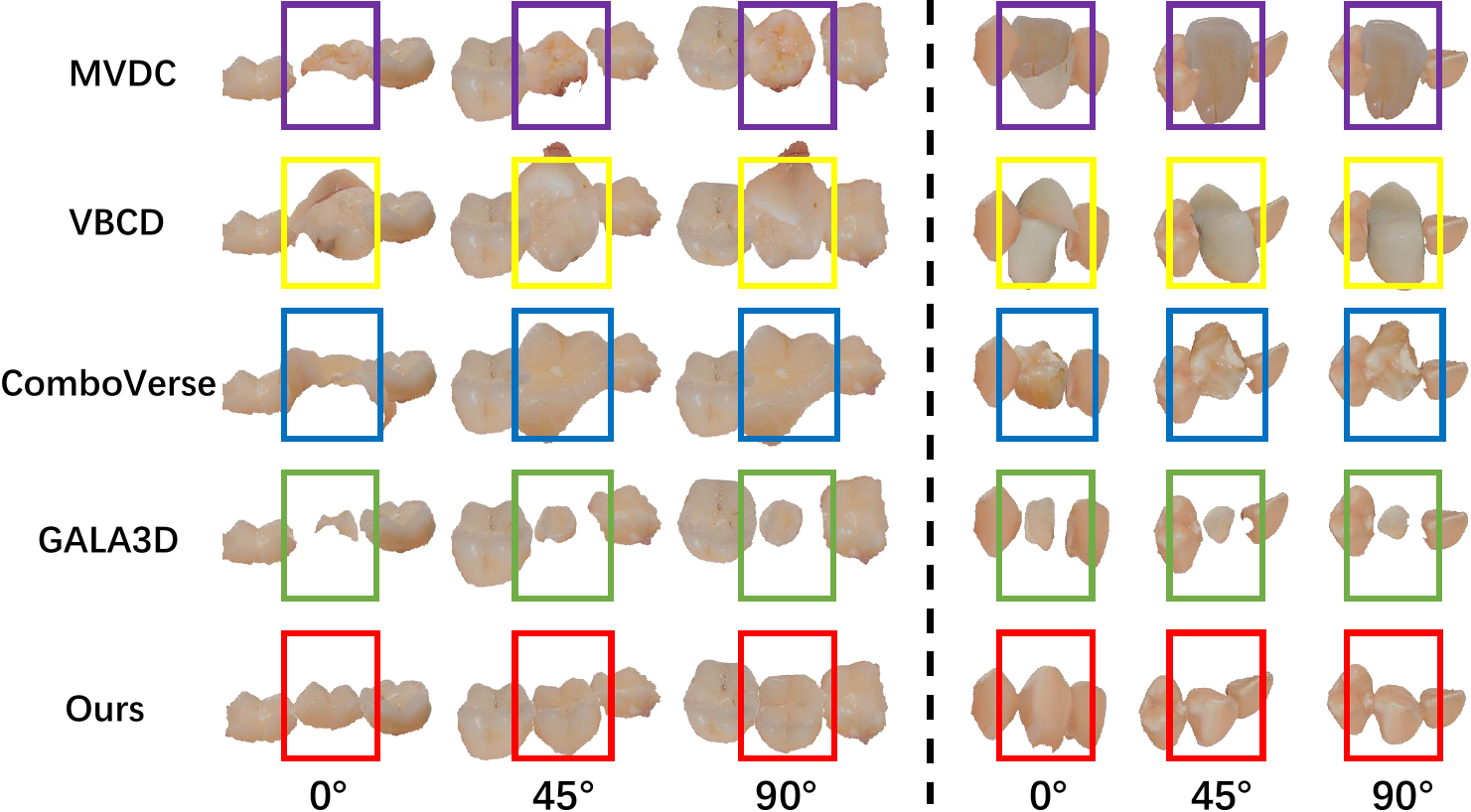} \\
\caption{\textbf{Qualitative comparisons of rendered images between MVDC, VBCD, ComboVerse, GALA3D, and our approach on the Shining3D tooth design dataset are illustrated.}
The results of the two samples are divided by the black dotted line.
Colorful bounding boxes are used for comparison of multiview consistency.}\label{figShining3D}
\end{figure*}

\begin{figure}[htbp]
\centering 
\includegraphics[width=0.99\linewidth]{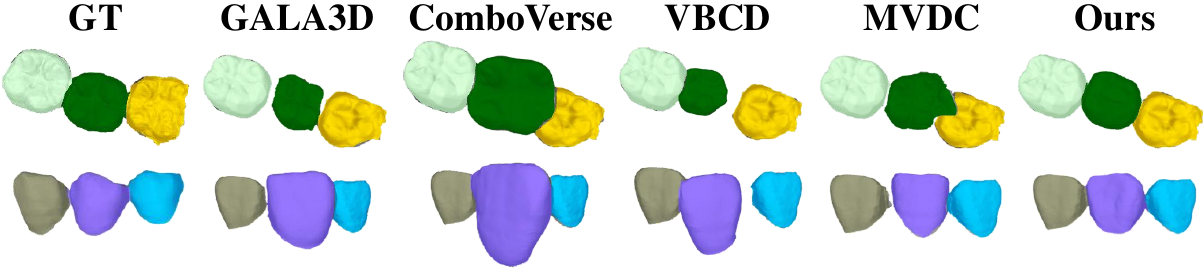} \\
\caption{\textbf{Qualitative comparisons of meshes between MVDC, VBCD, ComboVerse, GALA3D, and our approach on the Shining3D tooth design dataset are illustrated.}
Each row represents a specific sample. GT means the ground truth.}\label{figMesh}
\end{figure}

\textbf{Qualitative Comparison}.
Qualitative comparisons of rendered images among MVDC, VBCD, ComboVerse, GALA3D, and our approach using the Shining3D tooth design dataset are illustrated in Fig.~\ref{figShining3D}.
Our approach, which employs a graph diffusion model and a collision-loss mechanism based on 3D Gaussians, demonstrates the ability to generate multiple teeth that seamlessly integrate with the surrounding dentition, producing visually coherent and high-quality editing results.
Qualitative comparisons of the corresponding meshes among these approaches are illustrated in Fig.~\ref{figMesh}.
The results are consistent with those observed in the rendered-image comparisons: both the locations and surfaces of missing teeth are accurately predicted, thanks to our model’s awareness of global structure and local spatial relationships.

\begin{figure}[htbp]
\includegraphics[width=0.95\linewidth]{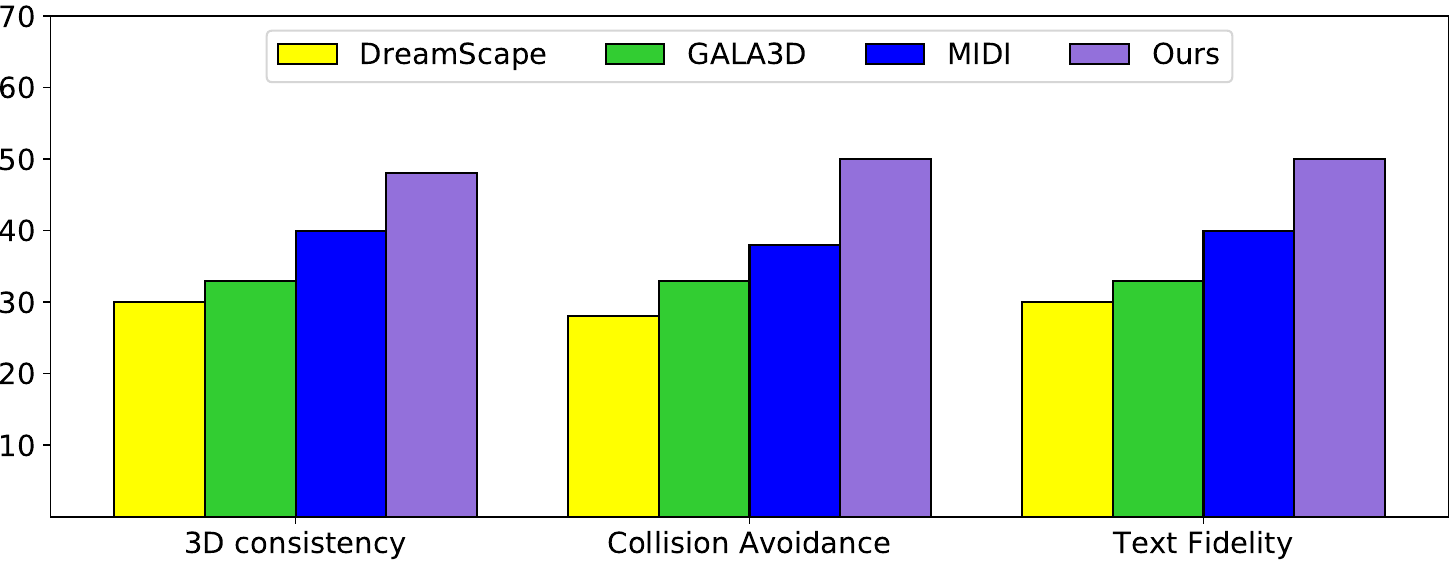}
\caption{\textbf{User Study.}
In a comprehensive user study encompassing three evaluation criteria that are 3D consistency, collision avoidance, and text fidelity, our approach achieves the highest scores of all the criteria.}\label{figUser}
\end{figure}

\textbf{User Study}.
To assess the subjective aspects of scene editing, we conducted a user study comparing our approach with state-of-the-art (SOTA) alternatives.
The study collected a total of 241 votes based on three primary criteria: 3D consistency, collision avoidance, and fidelity to textual descriptions.
As shown in Fig.~\ref{figUser}, our method was predominantly favored across these metrics.

\textbf{Resource Consumption}.
The proposed approach demonstrates a consistent reduction in instance conflicts across various settings, requiring an additional 0.4 GB of memory compared to GALA3D, with a training duration of one hour.
Furthermore, the efficiency comparison of different methodologies applied to the Shining3D dataset is presented in Table~\ref{tab_efficiency}.
Although the proposed method exhibits slower performance relative to ComboVerse and GALA3D, this additional computational overhead is justified by a 3.3-point improvement in FID over GALA3D and by the generation of collision-free geometry, making it suitable for precision-critical applications in dentistry.
\begin{table}[htbp]
  \centering
  \caption{\textbf{Efficiency comparison of different approaches in the Shining3D dataset.}
  }
  \label{tab_efficiency}
  \footnotesize 
  \begin{tabular}{cccccc}
\toprule
Approach &MVDream &GALA3D &ComboVerse &Ours\\
\midrule
Time (minute)  &\cellcolor{blue!10}\textbf{2.5}  &4.2  &4.4  &4.7\\
\bottomrule
\end{tabular}
\end{table}

\subsection{Discussion}
We propose a compositional 3D generation approach for oral scenarios that predicts the layout using a graph diffusion model.
Subsequently, optimizations are performed iteratively at both the scene and instance levels.
Among these optimizations, we introduce a collision-loss function based on 3D Gaussians to penalize tooth intersections, thereby enabling the simulation of multiple missing teeth when only text and image prompts are provided.

The experimental results derived from two public datasets illustrate the efficacy of the proposed approach.
Specifically, the layout prior improves significantly in compositional instance editing, as the spatial distribution of multiple neighboring elements is considered simultaneously during inference.
In addition, collision conflicts are mitigated by incorporating a shape prior based on intravariance within the 3D Gaussian splatting representation.

In our experiments, the average intravariance $R$ for typical teeth ranges from 3.0 mm to 6.0 mm, depending on the size of the tooth (e.g., molars exhibit larger $R$ values).
Assume that $h$ is the distance between the 3D Gaussian points of neighboring teeth and the center of the anchor tooth, the collision loss function effectively resolves overlaps when $h < R$, with a tolerance of approximately 0.1–0.3 mm—well within the clinical requirements for dental models.
For example, if $R = 3.0$ mm and $h = 2.8$ mm, the loss penalizes overlaps of 0.2 mm or greater.
In cases where $h > R$, minor overlaps may persist, but our dual-level optimization (Algorithm~\ref{alg:algorithm2}) ensures that such residuals are minimized through global scene consistency.

The efficacy of layout optimization via graph diffusion lies in its ability to model intricate structural dependencies and the anatomical constraints inherent in dental arrangements.
Unlike discrete graph generation, the diffusion process systematically denoises the target layout while integrating text–graph joint conditioning through cross-attention layers.
This methodology ensures semantic alignment with clinical requirements (e.g., tooth types) and geometric coherence (e.g., symmetry and occlusion).
Regularization based on KL divergence enforces consistency between the denoised and original jaw graphs, thereby maintaining biomechanical plausibility.
By conceptualizing teeth as interconnected nodes with hierarchical relationships, the model adeptly resolves multiscale spatial conflicts, thereby avoiding the local minima that are common in pairwise relation–based approaches.
This iterative refinement achieves a harmonious balance between global scene consistency and instance-level realism, as demonstrated by the improved FID and LPIPS metrics.
\begin{figure}[htbp]
  \centering
  \includegraphics[width=0.9\linewidth]{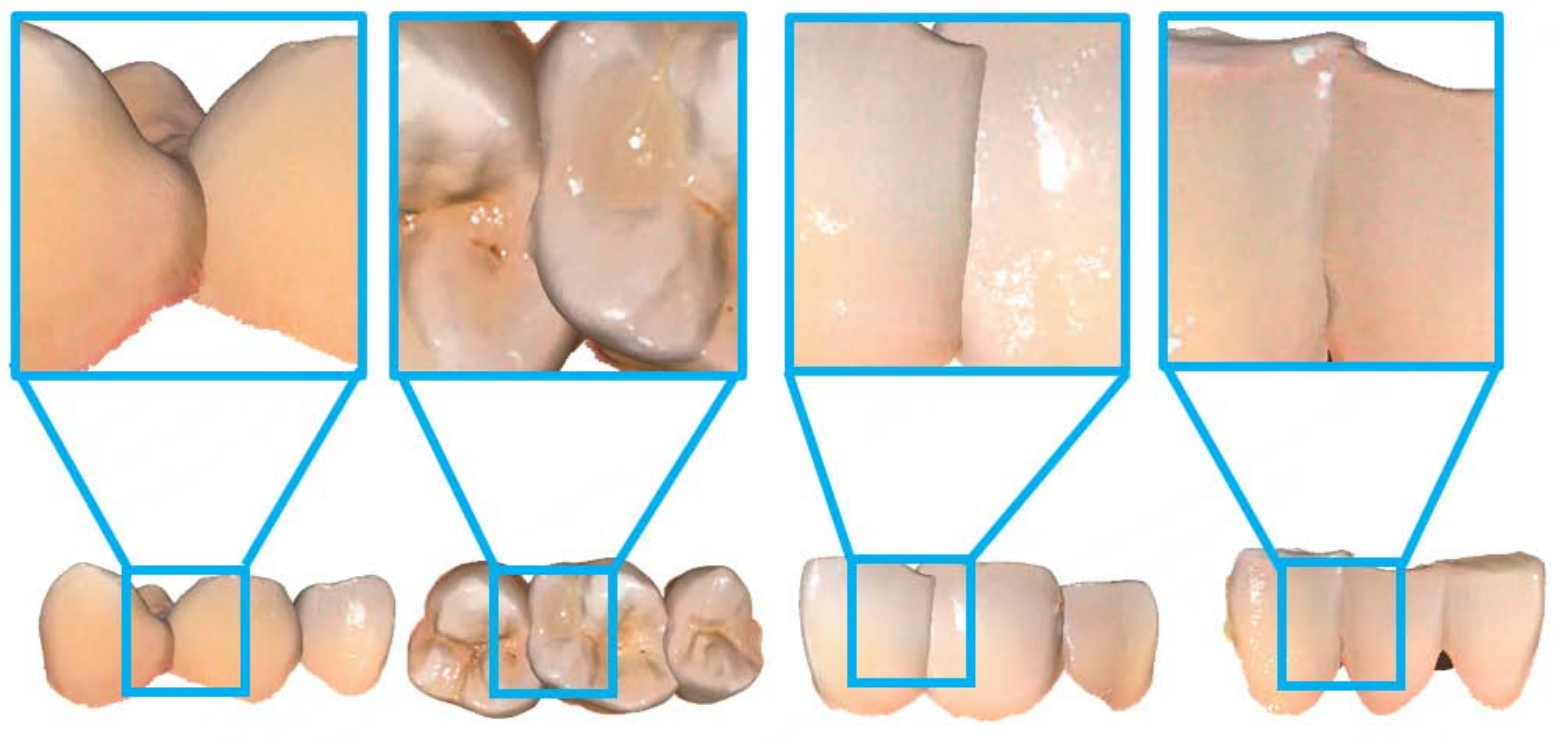}
  \caption{\textbf{Illustration of the drawbacks in our approach.}
  Multiple generated teeth may adhere to neighboring teeth.}
  \label{fig_drawback}
\end{figure}

Although our approach enhances the quality and robustness of compositional 3D tooth models, some limitations in generation remain to be addressed.
In compositional optimization for scenes and instances, the generated teeth may adhere to neighboring teeth due to their high similarity in appearance.
Examples are illustrated in Fig.~\ref{fig_drawback}.
The layout prior partially mitigates this issue by integrating biological knowledge and contextual information related to the jaw model; however, the problem of tooth adherence between adjacent teeth still persists.
Moreover, the efficiency of inference requires improvement.
Our approach requires approximately five minutes to generate multiple teeth, incorporating layout optimization through graph diffusion and compositional optimization for both scenes and instances.

For teeth with severe malformations, atypical implants, or pathological geometries, the symmetric intravariance $R_i$ (derived from Gaussian sparsity) becomes an unreliable collision threshold.
Instead, adaptive collision modeling should be introduced:
1) Surface-Aware Metrics: Replace centroid-based $R_i$ with curvature-aware or boundary-focused distances.
Extract mesh surfaces from Gaussians via lightweight Poisson reconstruction, then compute pointwise SDFs or nearest-edge distances for collision zones.
2) Hierarchical intravariance: Decompose irregular teeth into subregions (e.g., crown/root) and compute localized values of $R_i$. Apply the collision loss per subregion to handle asymmetric shapes.
To decrease inference time in compositional 3D tooth generation, a coarse-to-fine optimization strategy may be implemented.
Additionally, progressive Gaussian pruning could remove low-opacity splats during early optimization stages, reducing computational overhead in later refinements.

\section{Conclusion}  \label{conclusion}
In this paper, we present an approach for the compositional generation of 3D teeth.
This approach infers the tooth layout by progressively denoising the source graph using text and graph constraints.
Additionally, collision conflicts are mitigated by integrating a tooth shape prior based on 3D Gaussian splatting.
Comprehensive experiments from multiple datasets demonstrated that our approach consistently outperforms state-of-the-art approaches, including MVDC, VBCD, ComboVerse, and GALA3D, in terms of 3D consistency, collision prevention, and text fidelity. 
The integration of a dual-level optimization scheme further ensures global scene stability while minimizing local overlaps within clinically acceptable precision tolerances.

Despite these advancements, certain challenges remain. 
In particular, compositional optimization may occasionally cause adjacent teeth to adhere due to high inter-tooth similarity. 
Future research will explore refined biological priors, instance-level disentanglement strategies, and hybrid diffusion-neural field representations to further enhance structural differentiation and realism in complex dental reconstructions.

\bibliographystyle{IEEEtran}
\bibliography{mybibfile1}

\begin{IEEEbiography}[{\includegraphics[width=1.2in,height=1.2in,clip,keepaspectratio]{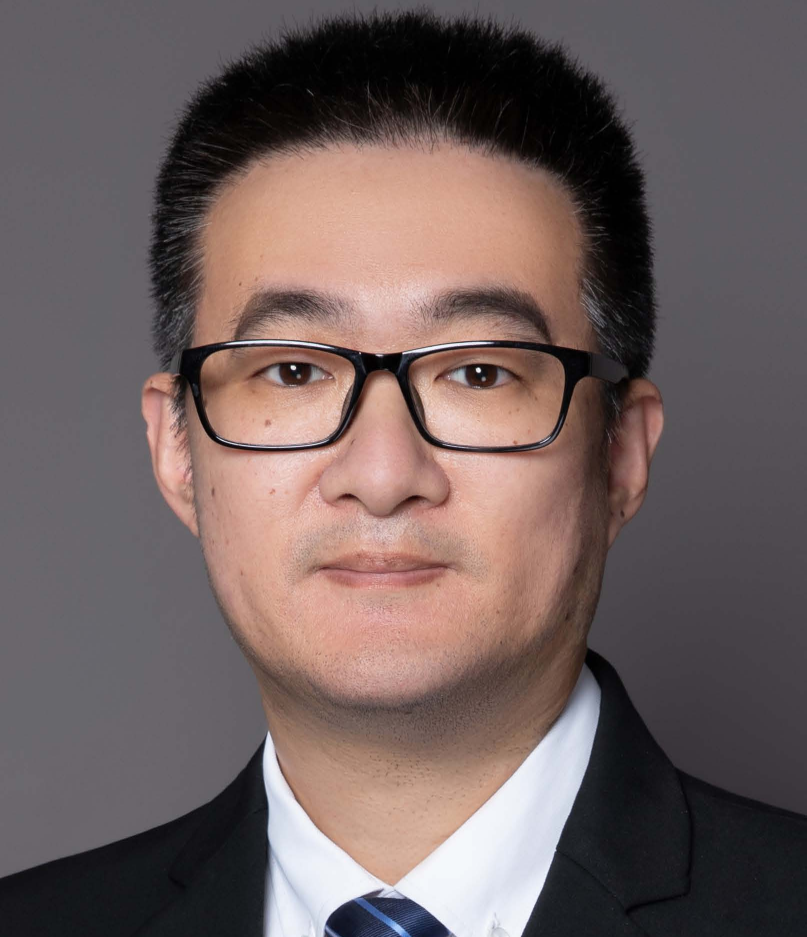}}]{Yan Tian}
received his Ph.D. degree from Beijing University of Posts and Telecommunications, Beijing, China, in 2011. Then he held a postdoctoral research fellow position (2012-2015) in the Department of Information and Electronic Engineering, Zhejiang University, Hangzhou, China. He is currently a Professor at the School of Computer Science and Technology of Zhejiang Gongshang University, China. His current interests are machine learning and video analysis.
\end{IEEEbiography}

\begin{IEEEbiography}[{\includegraphics[width=1.2in,height=1.2in,clip,keepaspectratio]{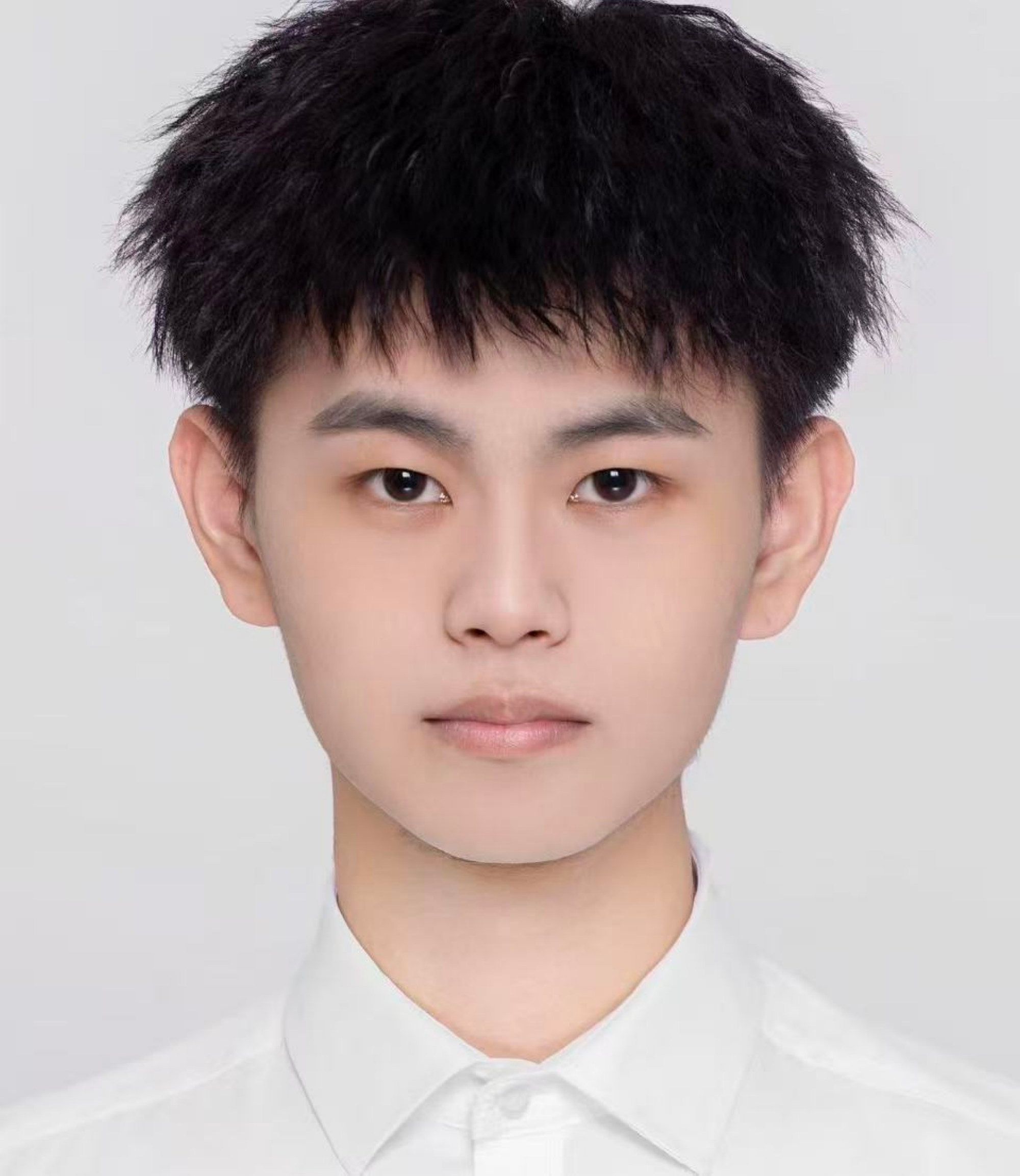}}]{Pengcheng Xue}
received his bachelor's degree from the School of Computer Science and Technology, Changchun University of Finance and Economics, China, in 2022. He is currently pursuing his Master's degree at the School of Computer Science and Technology, Zhejiang Gongshang University, China. His research interests include machine learning and computer vision.
\end{IEEEbiography}

\begin{IEEEbiography}[{\includegraphics[width=1.2in,height=1.2in,clip,keepaspectratio]{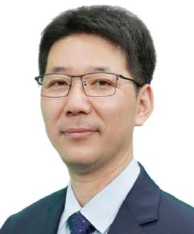}}]{Weiping Ding}
received the Ph.D. degree in Computer Science, Nanjing University of Aeronautics and Astronautics, Nanjing, China, in 2013. From 2014 to 2015, he was a Postdoctoral Researcher at the Brain Research Center, National Chiao Tung University, Hsinchu, Taiwan, China. In 2016, he was a Visiting Scholar at National University of Singapore, Singapore. From 2017 to 2018, he was a Visiting Professor at University of Technology Sydney, Australia. Now he is the Full Professor of Nantong University. His research directions involve granular data mining and multimodal machine learning. 
He serves as an Associate Editor/Area Editor/Editorial Board member of more than 10 international prestigious journals, such as IEEE Transactions on Neural Networks and Learning Systems, IEEE Transactions on Fuzzy Systems, IEEE/CAA Journal of Automatica Sinica, IEEE Transactions on Emerging Topics in Computational Intelligence, IEEE Transactions on Intelligent Transportation Systems, Information Fusion, Neurocomputing, Applied Soft Computing, et al. He was the Leading Guest Editor of Special Issues in several prestigious journals, including IEEE Transactions on Evolutionary Computation, IEEE Transactions on Fuzzy Systems, Information Fusion, et al.
\end{IEEEbiography}

\begin{IEEEbiography}[{\includegraphics[width=1.2in,height=1.2in,clip,keepaspectratio]{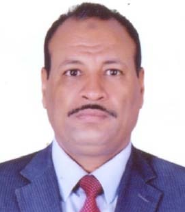}}]{Mahmoud Hassaballah}
received the Doctor of Engineering in Computer Science from Ehime University, Japan in 2011. He is currently a Professor of Computer Science at the Department of Computer Science, Prince Sattam Bin Abdulaziz University, Saudi Arabia. Also, he is a full Professor at the Department of Computer Science, Qena University, Egypt. He serves as a reviewer for several Journals such as IEEE Transactions on Image Processing, IEEE Transactions on Circuits and Systems for Video Technology, IEEE Transactions on Industrial Informatics, IEEE Transactions on Fuzzy Systems. Also, he is a TPC member of many conferences. He is an Editorial Board member of Pattern Analysis and Applications, Real-Time Image Processing, IET Image Processing, and Imaging Science Journal. His research interests include human-centered artificial intelligence, machine learning, computer vision, biometrics, image processing, feature extraction, object detection/recognition, and data security.
\end{IEEEbiography}

\begin{IEEEbiography}[{\includegraphics[width=1.2in,height=1.2in,clip,keepaspectratio]{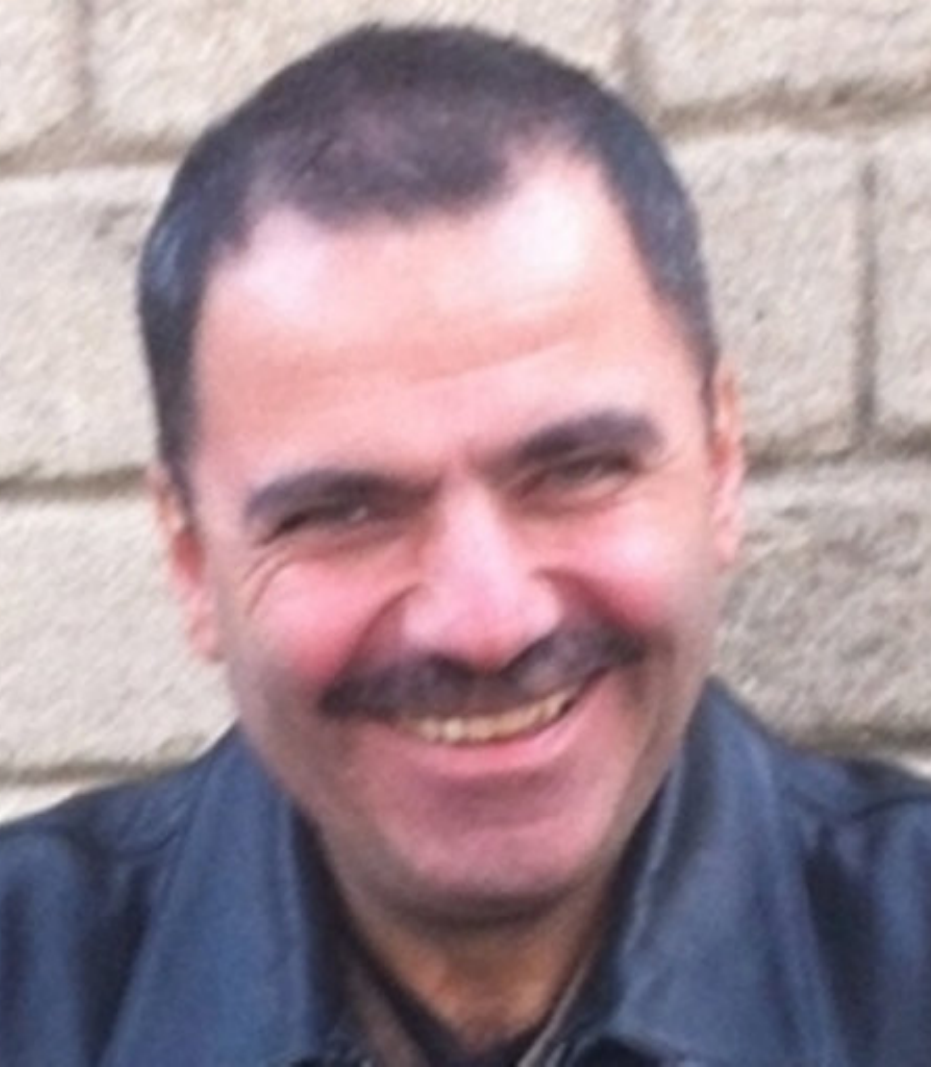}}]{Karen Egiazarian}
received Ph.D. degree in physics and mathematics from Moscow State University, Russia, in 1986, and Doctor of Technology in signal processing from Tampere University of Technology, Finland, in 1994. He is a Professor of Signal Processing at the Department of Computing Sciences, Tampere University, Tampere, Finland. He is an IEEE Fellow. He was Editor-in-Chief of the Journal of Electronic Imaging, served as associate editor of the IEEE Transactions on Image Processing. His main research interests are in the field of computational imaging, compressed sensing, efficient signal processing algorithms, image/video restoration and compression.
\end{IEEEbiography}

\begin{IEEEbiography}[{\includegraphics[width=1.2in,height=1.2in,clip,keepaspectratio]{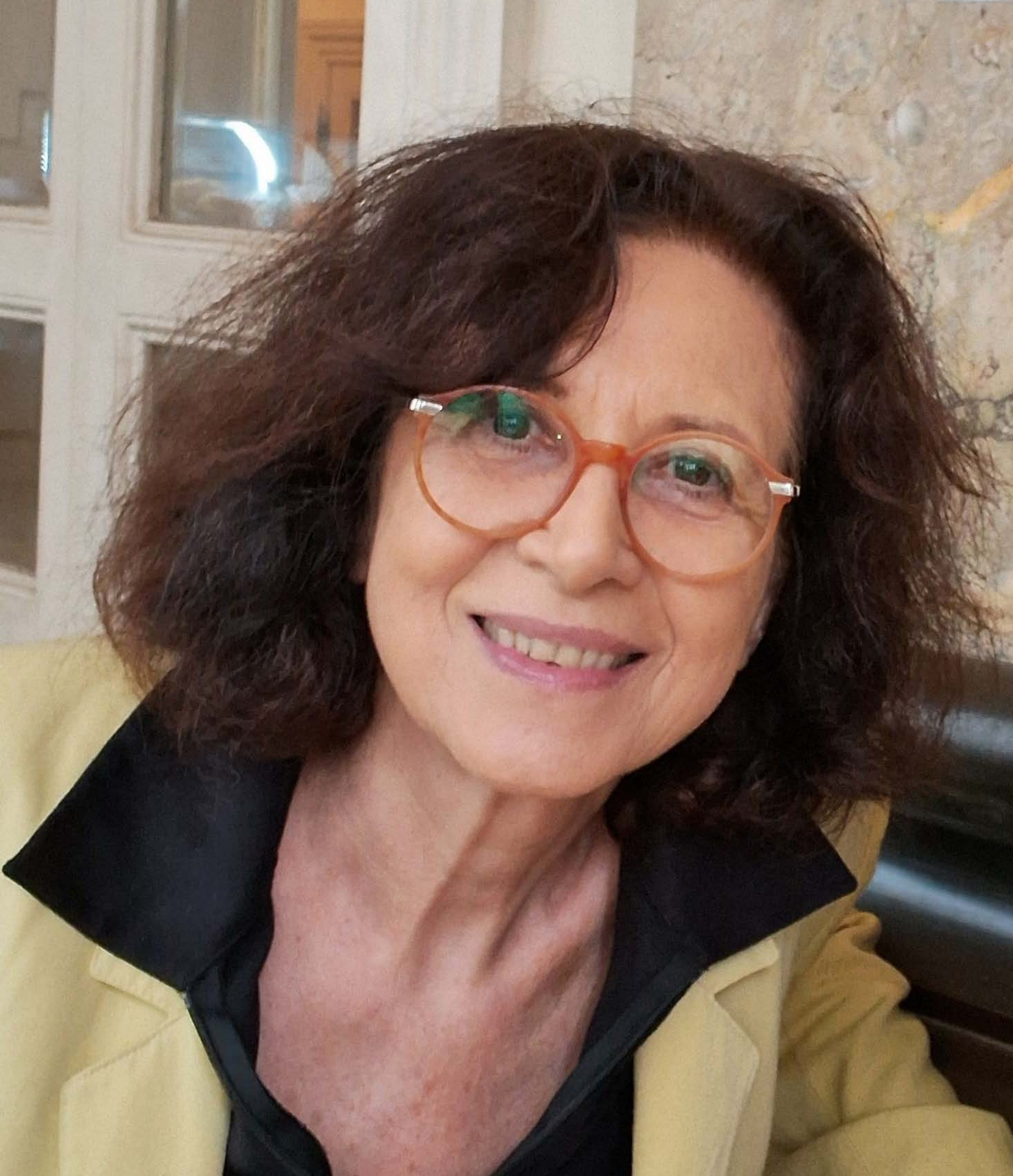}}]{Aura Conci}
is an engineer with M.Sc. and Ph.D. in structures, professor at Universidade Federal Fluminense (UFF). She works now in the areas of computer modeling, computer vision, image analysis and bioinformatics. She oriented hundred students and is a member of the: ACM, ISGG, ABCM, ISPRS, SBrT and SBC. She acts in the editorial office of a number of international journals and has cooperated on research with many scholars in various countries. She has a number of high quality publications (around 6k citations, h index=41 and i10=120). She has funding from Brazilian and EU governments for coordinating over 30 research projects.
\end{IEEEbiography}

\begin{IEEEbiography}[{\includegraphics[width=1.2in,height=1.2in,clip,keepaspectratio]{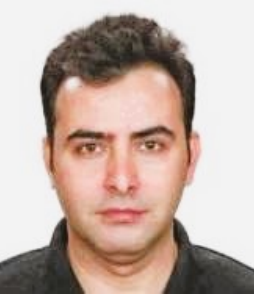}}]{Abdulkadir Sengur}
received the B.Sc. degree in electronics and computers education, the M.Sc. degree in electronics education, and the Ph.D. degree in electrical and electronics engineering from Firat University, Turkey, in 1999, 2003, and 2006, respectively. He became a Research Assistant with the Technical Education Faculty, Firat University, in February 2001. He is currently a Professor with the Technology Faculty, Firat University. His research interests include signal processing, image segmentation, pattern recognition, medical image processing, and computer vision.
\end{IEEEbiography}

\begin{IEEEbiography}[{\includegraphics[width=1.2in,height=1.2in,clip,keepaspectratio]{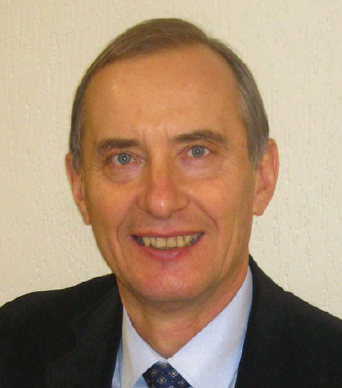}}]{Leszek Rutkowski}
received the M.Sc., Ph.D., and D.Sc. degrees from the Wrocław University of Technology, Wrocław, Poland, in 1977, 1980, and 1986, respectively, and the Honoris Causa degree from the AGH University of Science and Technology, Krak´ow, Poland, in 2014. He is with the Systems Research Institute of the Polish Academy of Sciences, Warsaw, Poland, and with the Institute of Computer Science, AGH University of Science and Technology, Krakow, Poland, in both places serving as a professor. He is an Honorary Professor of the Czestochowa University of Technology, Poland, and he also cooperates with the University of Social Sciences in Ł´od´z, Poland. His research interests include machine learning, data stream mining, big data analysis, neural networks, stochastic optimization and control, agent systems, fuzzy systems, image processing, pattern classification, and expert systems. He has published seven monographs and more than 300 technical papers, including more than 40 in various series of IEEE Transactions. He is the president and founder of the Polish Neural Networks Society. He is on the editorial board of several most prestigious international journals. He is a recipient of the IEEE Transactions on Neural Networks Outstanding Paper Award. He is a Full Member (Academician) of the Polish Academy of Sciences, elected in 2016, and a Member of the Academia Europaea, elected in 2022. He is also a Life Fellow of IEEE.
\end{IEEEbiography}

\end{document}